\documentclass[pdflatex,sn-mathphys-num]{sn-jnl}

\usepackage{graphicx}%
\usepackage{subcaption}%
\usepackage{float}%
\usepackage{multirow}%
\usepackage{amsmath,amssymb,amsfonts}%
\usepackage{amsthm}%
\usepackage{mathrsfs}%
\usepackage[title]{appendix}%
\usepackage{xcolor}%
\usepackage{textcomp}%
\usepackage{manyfoot}%
\usepackage{booktabs}%
\usepackage{algorithm}%
\usepackage{algorithmicx}%
\usepackage{algpseudocode}%
\usepackage{listings}%
\usepackage{rotating}   
\usepackage[referable]{threeparttablex}

\usepackage{longtable}
\usepackage{booktabs}
\usepackage{multirow}
\usepackage{rotating}
\usepackage{svg}

\theoremstyle{thmstyleone}%

\theoremstyle{thmstyletwo}%

\theoremstyle{thmstylethree}%

\algrenewcommand\algorithmicrequire{\textbf{Input:}}
\algrenewcommand\algorithmicensure{\textbf{Output:}}

\begin{document}

\title[MSDE]{Anomaly Detection via Mean Shift Density Enhancement}


\author[1]{\fnm{Pritam} \sur{Kar}}\email{karpritam25@iisertvm.ac.in}\equalcont{These authors contributed equally to this work}

\author[2]{\fnm{Rahul} \sur{Bordoloi}}\email{rahul.bordoloi@uni-rostock.de}\equalcont{These authors contributed equally to this work}

\author[2,3,4]{\fnm{Olaf} \sur{Wolkenhauer}}\email{olaf.wolkenhauer@uni-rostock.de}

\author*[1]{\fnm{Saptarshi} \sur{Bej}}\email{sbej7042@iisertvm.ac.in}

\affil*[1]{\orgdiv{School of Data Science}, \orgname{Indian Institute of Science Education and Research}, \orgaddress{\city{Thiruvananthapuram}, \postcode{695551}, \state{Kerala}, \country{India}}}

\affil[2]{\orgdiv{Institute of Computer Science}, \orgname{University of Rostock}, \orgaddress{\city{Rostock}, \postcode{18051}, \country{Germany}}}

\affil[3]{\orgdiv{Leibniz-Institute for Food Systems Biology}, \orgname{Technical University of Munich}, \orgaddress{\state{Freising}, \country{Germany}}}

\affil[4]{\orgname{Stellenbosch Institute of Advanced Studies (STIAS)}, \orgaddress{\country{South Africa}}}



\abstract{
Unsupervised anomaly detection stands as an important problem in machine learning. Existing unsupervised anomaly detection algorithms rarely perform well across different anomaly types, often excelling only under specific structural assumptions. This lack of robustness also becomes particularly evident under noisy settings. 

\quad We propose \emph{Mean Shift Density Enhancement} (MSDE), a fully unsupervised framework that detects anomalies through their \emph{geometric response} to density-driven manifold evolution. MSDE is designed as a general purpose anomaly detection framework, based on the principle that normal samples, being well supported by local density, remain stable under iterative density enhancement, whereas anomalous samples undergo large cumulative displacements as they are attracted toward nearby density modes. To operationalize this idea, MSDE employs a weighted mean-shift procedure with adaptive, sample-specific density weights derived from a manifold learning-based fuzzy neighborhood graph.

\quad We evaluate MSDE on an anomaly detection benchmark comprising 46 real-world tabular datasets, four realistic anomaly generation mechanisms, and six noise levels. Compared to 13 established unsupervised baselines, MSDE achieves consistently strong, balanced and robust performance for several standard classification metrics, at several noise levels and on average over several types of anomalies. These results demonstrate that displacement-based scoring provides a robust alternative to the existing state-of-the-art for unsupervised anomaly detection.
}




\maketitle

\section{Introduction}\label{sec:intro}

Anomaly detection is a fundamental problem in machine learning with applications in fraud detection, network intrusion monitoring, medical diagnostics, and several other fields. In many real-world settings, anomalies do not simply manifest as isolated outliers in Euclidean space; instead, they appear as subtle deviations from the underlying structure of a complex data manifold.

A framework called the Anomaly Detection Benchmark (ADBench)~\citep{jiang2022adbench} organizes anomalies into four distinct types based on the scale and structure in which they occur. These are: anomalies that are isolated from the global distribution, anomalies that only deviate within a local context, clusters of anomalous samples that exhibit similar characteristics but are nevertheless distinguishable from the general group distribution, and anomalies that devaite from the dependency structure of normal data. A summary of this taxonomy is provide in Table~\ref{tab:adbench_anomaly_types}


\begin{table}[htbp!]
\centering
\caption{Overview of anomaly types used in the ADBench benchmark.}
\label{tab:adbench_anomaly_types}
\begin{tabular}{p{3cm} p{9.5cm}}
\toprule
\textbf{Anomaly Type} & \textbf{Description} \\
\midrule
Global anomalies &
Samples that deviate significantly from the overall data distribution and are isolated in the feature space. These anomalies are typically far from dense regions and are often detectable by global distance- or density-based methods. \\

Local anomalies &
Samples that appear normal with respect to the global distribution but are anomalous relative to their local neighborhood. These anomalies are embedded within dense regions yet exhibit subtle deviations from local structure, making them challenging for global detectors. \\

Cluster anomalies &
Small or sparse clusters of samples that are internally coherent but differ from dominant clusters in the dataset. Such anomalies violate assumptions of cluster size or density and are often difficult to detect using point-wise outlier criteria alone. \\

Dependency anomalies &
Samples that individually appear normal in marginal feature distributions but violate dependencies or correlations between features. These anomalies reflect structural inconsistencies rather than marginal extremeness and require detectors sensitive to multivariate relationships. \\
\bottomrule
\end{tabular}
\end{table}

Despite substantial methodological diversity as shown in Section~\ref{sec:relatedWork}, ADBench demonstrates that no unsupervised anomaly detection method achieves consistently robust performance across datasets, anomaly types, and noise regimes~\citep{jiang2022adbench}. Our experimental results (Section~\ref{sec:results}, and Table~\ref{tab:syntheticmode_multimodel} in Appendix~\ref{secA2}) further support this observation: different classes of methods are effective for different anomaly structures. For instance, local density and nearest-neighbor approaches are particularly effective for detecting local anomalies, whereas projection- and variance-based methods more reliably identify clustered anomalies. Nevertheless, ADBench reports that no unsupervised approach is statistically superior overall.

These findings motivate the exploration of methods that extend beyond modeling a static data distribution. Instead, we consider iterative procedures that progressively transform samples toward the underlying data distribution, while identifying as anomalous those samples that require the greatest degree of transformation to ``regularise.'' The underlying intuition is that anomaly detection should not depend on a single predefined metric, but rather on the extent to which a sample must be modified to conform to the dominant structure of the data distribution.

\subsection{Our contribution}

We propose \emph{Mean Shift Density Enhancement} (MSDE), a non-parametric anomaly detection framework that exploits local geometric structure through an adaptive mean-shift mechanism. MSDE is motivated by two key observations: (i) anomalies tend to lie in regions of low or unstable local density, and (ii) iterative shifting toward locally estimated density modes naturally exposes such irregularities. 

Unlike classical mean-shift methods, MSDE constructs data-dependent weights using a fuzzy neighborhood graph derived from the Uniform Manifold Approximation and Projection (UMAP)~\citep{mcinnes2018umap} algorithm, enabling adaptation to heterogeneous manifolds and arbitrary data distributions. MSDE first builds an adaptive neighborhood graph, then estimates multi-scale, sample-specific density weights, and finally performs a weighted mean-shift procedure. The cumulative movement of each point across iterations serves as its anomaly score: points embedded deeply within the manifold shift minimally, whereas anomalous points undergo large, directed movement toward density modes. Our contributions are as follows:
\begin{enumerate}
    \item \textbf{A displacement-based principle for anomaly detection.}
    We introduce cumulative geometric shift as a novel anomaly signal, moving beyond static density or distance measures. This criterion provides a principled and interpretable mechanism for distinguishing normal and anomalous samples based on their stability under iterative density enhancement.

    \item \textbf{A manifold-adaptive mean-shift framework.}  
    We develop a weighted mean-shift procedure in which neighborhood influence is governed by empirical density estimates derived from a UMAP-based fuzzy neighborhood graph. This enables MSDE to adapt naturally to heterogeneous  data manifolds without explicit parametric assumptions.

\end{enumerate}

Through extensive evaluation on the ADBench benchmark, we demonstrate that MSDE achieves consistently strong performance across diverse anomaly types and noise levels, outperforming $13$ established unsupervised baselines on an average over different noise levels and anomaly types, even with a default parameter setting. MSDE thus offers a principled, manifold-aware approach to anomaly detection.

\subsection{Related Work}\label{sec:relatedWork}

Anomaly detection methods in the literature broadly fall into three categories: supervised, semi-superised, and unsupervised approaches, which are summarised in Table~\ref{tab:ad_taxonomy}.

\paragraph{Supervised anomaly detection.}
In the fully supervised regime, anomaly detection reduces to binary classification, where labeled anomalies define discriminative decision boundaries (Table~\ref{tab:ad_taxonomy}). Tree-based ensembles and gradient-boosted models such as LightGBM (2017)~\citep{ke2017lightgbm}, XGBoost (2016)~\citep{chen2016xgboost}, and CatBoost (2018)~\citep{prokhorenkova2018catboost}, as well as neural tabular architectures including ResNet-style models (2016)~\citep{he2016resnet} and FT-Transformer (2021)~\citep{gorishniy2021fttransformer}, achieve strong performance when anomaly labels are abundant and representative. However, their reliance on labeled anomalies makes them vulnerable to label scarcity, selection bias, and shifts in anomaly semantics~\citep{pang2021adreview}.

\paragraph{Semi-supervised anomaly detection.}
Semi-supervised methods exploit a limited number of anomaly labels to guide representation learning or score calibration (Table~\ref{tab:ad_taxonomy}). Representative approaches include deviation-based learning (DevNet) (2019)~\citep{pang2019deep}, margin-based objectives (DeepSAD) (2020)~\citep{ruff2020deep}, and neighborhood-aware refinements such as PReNet (2019) and FEAWAD (2021)~\citep{ren2019progressive,Zhou_Song_Zhang_Liu_Zhu_Liu_2021}. Hybrid models like XGBOD (2018)~\citep{zhao2018xgbod} integrate unsupervised signals into supervised learners. While semi-supervised anomaly detection (SSAD) (2025)~\citep{Yoo_Zhao_Akoglu_2025} improves robustness over fully supervised setups, performance remains sensitive to label noise and distributional drift.

\paragraph{Unsupervised anomaly detection.}
In the absence of labeled anomalies, unsupervised anomaly detection dominates real-world deployments. Under the ADBench formulation, given data $X={x_i}_{i=1}^n$ drawn from a mixture of mostly normal samples and a small fraction of anomalies, the goal is to assign anomaly scores $s(x_i)$ based on structural assumptions about normality. These assumptions include geometric compactness (DeepSVDD (2018)~\citep{ruff2018deepsvdd}), local density consistency (LOF (2000)~\citep{breunig2000lof}), distributional regularity (ECOD (2022)~\citep{li2022ecod}), and stability under random projections or reconstruction (OCSVM (2001)~\citep{scholkopf2001ocsvm}).

\begin{table}
\centering
\caption{Expanded taxonomy of anomaly detection paradigms, their modeling assumptions, and representative methods.}
\label{tab:ad_taxonomy}
\small
\begin{tabular}{p{2.2cm} p{3.2cm} p{3.2cm} p{3.0cm}}
\toprule
\textbf{Paradigm / Family} &
\textbf{Core Assumption / Signal} &
\textbf{Scoring Mechanism} &
\textbf{Representative Methods} \\
\midrule

Supervised AD &
Labeled anomalies define decision boundary &
Classification margin or posterior probability &
LightGBM~\citep{ke2017lightgbm}, XGBoost~\citep{chen2016xgboost}, CatBoost~\citep{prokhorenkova2018catboost}; tabular MLP, ResNet~\citep{he2016resnet}, FT-Transformer~\citep{gorishniy2021fttransformer} \\

Semi-supervised AD &
Few anomaly labels refine representations or scores &
Deviation, margin, or label-guided ranking &
DevNet~\citep{pang2019deep}, DeepSAD~\citep{ruff2020deep}, PReNet~\citep{ren2019progressive}, FEAWAD~\citep{Zhou_Song_Zhang_Liu_Zhu_Liu_2021}, XGBOD~\citep{zhao2018xgbod}, SSAD~\citep{Yoo_Zhao_Akoglu_2025} \\

Projection-based &
Normal data lie in low-dimensional subspace &
Reconstruction error or projection residual &
PCA~\citep{shyu2003pca}, OCSVM~\citep{scholkopf2001ocsvm} \\

Distance-based &
Anomalies occur in sparse regions &
$k$-NN distance or rank statistics &
KNN detector~\citep{ramaswamy2000knn} \\

Local density-based &
Anomalies deviate from neighbor-relative density &
Density ratio or cluster-weighted score &
LOF~\citep{breunig2000lof}, CBLOF~\citep{he2003cblof}, COF~\citep{tang2002cof}, SOD~\citep{kriegel2009sod} \\

Distributional / statistical &
Feature-wise or joint tail events &
Histogram, copula, or ECDF tails &
HBOS~\citep{goldstein2012hbos}, COPOD~\citep{li2020copod}, ECOD~\citep{li2022ecod} \\

Ensemble AD &
Randomized partitions improve robustness &
Average isolation depth or density &
Isolation Forest~\citep{liu2008isolation}, LODA~\citep{pevny_loda_2016} \\

Deep unsupervised &
Normal data form compact nonlinear manifold &
Latent hypersphere or density estimate &
DeepSVDD~\citep{ruff2018deepsvdd}, DAGMM~\citep{zong2018dagmm} \\

\bottomrule
\end{tabular}
\end{table}

\section{Mean Shift Density Enhancement}\label{sec:msde}

We propose \textbf{Mean Shift Density Enhancement (MSDE)}, a non-parametric anomaly detection method that detects anomalies by estimating the intrinsic structure of the data manifold through \emph{adaptive neighborhood weighting} (explained in Sections~\ref{sec:nbdGraph} and \ref{sec:nbdWeights}) and \emph{iterative shifting of data points} (explained in Section~\ref{sec:manifoldShift}). MSDE assigns anomaly scores by measuring the cumulative deviation or shift of each point from its locally estimated density mode across multiple shift iterations.

Our method consists of three components: (i) construction of an adaptive neighborhood graph, (ii) estimation of empirical sample weights via a UMAP-based fuzzy neighborhood graph, and (iii) a mean-shift procedure on the learned manifold.

\subsection{Adaptive Neighborhood Graph Construction}\label{sec:nbdGraph}

Let $X = \{x_i\}_{i=1}^N \subset \mathbb{R}^d$ be the dataset, $x_i$ being the rows of the data matrix. MSDE builds a local neighborhood structure using an approximate $k$-nearest neighbor graph constructed via NN-Descent in the current feature space. For each point $x_i$, we obtain
\begin{equation}
    \mathcal{N}_k(x_i) = \{x_{i_1}, \dots, x_{i_k}\},
\end{equation}
where $\mathcal{N}_k(x_i)$ denotes the $k$ nearest neighbors of $x_i$ identified through Euclidean distance queries using the NN-Descent algorithm. This neighborhood graph is recomputed at every mean-shift iteration to ensure that locality is defined with respect to the current estimate of the data manifold. Next, MSDE assigns \emph{sample-specific weights} that reflect local density and neighborhood reliability.

\subsection{Estimation of Empirical Neighborhood Weights}\label{sec:nbdWeights}

To construct robust, locality-aware weights, we compute a \emph{fuzzy neighborhood graph} using the UMAP framework. For each mini-batch of size $B$, we compute a fuzzy simplicial set
\begin{equation}
    G = \mathrm{UMAPGraph}(X_{\text{batch}}) \text{ (see Section 3.1 of \cite{mcinnes2018umap})},
\end{equation}
where $G \in [0,1]^{B \times B}$ contains pairwise membership strengths. Each entry $g_{ij}$ represents the membership strength between points $x_i$ and $x_j$ for indices $i,j \in \{1,\dots,B\}$. These values lie in the interval $[0,1]$ because the UMAP graph construction defines edge weights through a normalized exponential kernel and subsequent fuzzy set union operations, both of which produce values bounded between 0 and 1. Thus, $g_{ij}$ can be interpreted as the degree to which $x_i$ and $x_j$ belong to the same local manifold region.

\paragraph{Binary-search radius estimation.}
A key component of MSDE is a binary-search procedure that identifies a radius $\varepsilon$ for which at least a proportion $\alpha$ of the data points satisfy the required neighborhood-density condition.
\begin{equation}
    \#\{x_j : d_G(x_i, x_j) < \varepsilon\} > T_{\text{nbd}},
\end{equation}
where $d_G$ denotes distances in the similarity space derived from $G$, and $T_{\text{nbd}}$ is a predefined neighborhood-density threshold. 
If no radius satisfies the original density and proportion constraints, the conditions are automatically relaxed and the search is repeated; if still unsuccessful, the maximum observed distance is used as a fallback radius.

\paragraph{Averaged multi-radius weighting.}
We evaluate neighborhood densities at a sequence of decreasing radii
\begin{equation}
    \varepsilon_1 = \varepsilon,~
    \varepsilon_2 = \varepsilon - \Delta,~
    \dots,~
    \varepsilon_m = \varepsilon - (m-1)\Delta,
\end{equation}
and define the empirical weight for the $i$-th sample as
\begin{equation}
    w_i = \frac{1}{m} \sum_{r=1}^{m} 
    \#\{x_j : d_G(x_i, x_j) < \varepsilon_r\}.
\end{equation}
This yields smooth, stable weights capturing intrinsic density across multiple neighborhood scales.

Here, the decrement $\Delta$ is computed automatically based on the
radius $\varepsilon$ obtained from the binary–search step.
In the implementation, $\Delta$ is set as
\begin{equation}
\label{delta_calculation}
    \Delta = \frac{\varepsilon - 10^{-6}}{m},
\end{equation}
ensuring that the sequence of radii $\{\varepsilon_r\}_{r=1}^{m}$ decreases uniformly from the initial estimate $\varepsilon$ toward a small positive value. This produces a set of progressively stricter neighborhood scales, allowing the algorithm to average density estimates across multiple granularities and thereby obtain smoother and more stable empirical weights.

\subsection{Mean Shift on the Learned Manifold}\label{sec:manifoldShift}

Given empirical weights $\{w_j\}$ and local neighborhoods $\mathcal{N}_k(x_i)$ obtained via approximate nearest neighbor search, MSDE performs a \emph{weighted mean-shift update} on the learned manifold. At iteration $t$, we first compute a locally weighted neighborhood mean
\begin{equation}
    \mu_i^{(t)} = 
    \sum_{x_j \in \mathcal{N}_k(x_i)} 
    \tilde{w}_j x_j,
    \qquad 
    \tilde{w}_j = 
    \frac{w_j}{\sum_{x_\ell \in \mathcal{N}_k(x_i)} w_\ell}.
\end{equation}
We then update the position of $x_i$ along the direction from $x_i^{(t)}$ to $\mu_i^{(t)}$ with a normalized, magnitude-scaled step:
\begin{equation}
    x_i^{(t+1)} = x_i^{(t)} 
    + \eta \, \left\|x_i^{(t)} - \mu_i^{(t)}\right\| 
    \frac{\mu_i^{(t)} - x_i^{(t)}}{\left\| \mu_i^{(t)} - x_i^{(t)}\right\| + \epsilon},
\end{equation}
where $\eta$ is a learning rate and $\epsilon$ is a small constant for numerical stability. The instantaneous movement magnitude at iteration $t$ is defined as
\begin{equation}
    \delta_i^{(t)} = \left\| x_i^{(t+1)} - x_i^t \right\|.
\end{equation}
We accumulate the total distance traveled by each point across $T$ iterations:
\begin{equation}
    D_i = \sum_{t=1}^{T} \delta_i^{(t)}.
\end{equation}

\subsection{Anomaly Scoring}

Intuitively, \emph{normal samples} lie close to high-density regions of the manifold and remain relatively stable during the mean-shift iterations, resulting in small cumulative movement $D_i$. In contrast, \emph{anomalies}, being far from dense regions, are pulled significantly toward the manifold and therefore experience larger movement.

We normalize and transform the cumulative movement into anomaly scores using a sigmoid function
    $s_i = \sigma\!\big(\mathrm{scale}(D_i)\big),$
where $\mathrm{scale}(\cdot)$ denotes a standardization step (e.g., z-score normalization) and $\sigma$ is the logistic function. Higher scores correspond to more anomalous samples.

\paragraph{Stopping criterion.}
The mean-shift iterations are terminated using a simple convergence test. Let
\(\delta_i^{(t)} = \lVert x_i^{(t)} - \mu_i^{(t)} \rVert\) denote the movement of point $x_i$ at iteration $t$. We monitor the average movement
\begin{equation}
    \bar{\delta}^{(t)} = \frac{1}{N} \sum_{i=1}^N \delta_i^{(t)}.
\end{equation}
The algorithm stops either when a maximum number of iterations $T$ is reached or when the average movement falls below a small threshold $\tau$, $\bar{\delta}^{(t)} < \tau,$
indicating that the points have stabilized around the underlying density structure. The entire procedure is summarised in Algorithm~\ref{alg:msde_impl}.

\begin{table}[htbp!]
\centering
\caption{Main MSDE parameters with mathematical notation, code names, descriptions, and default values.}
\label{tab:msde_parameters}
\begin{tabular}{l l p{5.3cm} c}
\hline
\textbf{Notation} & \textbf{Code Name} & \textbf{Description} & \textbf{Default} \\
\hline
$k$ & \texttt{k} & Number of nearest neighbors used in each mean-shift update. & 100 \\
\hline
$T_{\text{nbd}}$ & \texttt{nbd\_sample\_count\_threshold} & Minimum number of neighbors required inside the estimated radius during weight computation. Controls density sensitivity. & 70 \\
\hline
$\eta$ & \texttt{learning\_rate} & Learning rate controlling the magnitude of each mean-shift update. & 0.1 \\
\hline
$T$ & \texttt{max\_iters\_shift} & Maximum number of mean-shift iterations. & 6 \\
\hline
$\tau$ & \texttt{shift\_threshold} & Stopping threshold: algorithm halts early when average movement drops below $\tau$. & 0.003 \\
\hline
$m$ & \texttt{max\_iters\_weight\_count} & Number of radii used when averaging density counts to compute empirical weights. & 4 \\
\hline
$\alpha$ & \texttt{satisfiability\_proportion} & Fraction of samples that must meet the density condition during binary-search estimation of $\varepsilon$. & 0.3 \\
\hline
\end{tabular}

\end{table}

\begin{algorithm}[htbp!]
\caption{Mean Shift Density Enhancement (MSDE)}
\label{alg:msde_impl}
\begin{algorithmic}[1]

\Require
Dataset $X = \{x_i\}_{i=1}^N \subset \mathbb{R}^d$,
number of neighbors $k$,
learning rate $\eta$,
maximum shift iterations $T$,
shift stopping threshold $\tau$,
neighborhood count threshold $T_{\text{nbd}}$,
number of radius levels $m$,
satisfiability proportion $\alpha$

\Ensure
Anomaly scores $\{s_i\}_{i=1}^N$

\vspace{0.5em}
\State \textbf{/* Step 1: Density-aware weight estimation */}

\State Partition $X$ into batches (or use full dataset if small)

\For{each batch $X_b$}
    \State Compute approximate $k$NN using NNDescent
    \State Construct UMAP fuzzy simplicial graph $G_b$
    \State Represent each point by its fuzzy similarity vector $g_i$
    
    \State Compute pairwise Euclidean distances between $\{g_i\}$
    \State Build KD-Tree on $\{g_i\}$

    \State Find radius $\varepsilon$ via binary search such that
    $\#\{g_j : \|g_i - g_j\| < \varepsilon\} > T_{\text{nbd}}$
    holds for at least $\alpha |X_b|$ points

    \For{$r = 1$ to $m$}
        \State $\varepsilon_r \gets \varepsilon - (r-1)\Delta$ (see equation \ref{delta_calculation})
        \State Count neighbors within radius $\varepsilon_r$
    \EndFor

    \State Assign weight
    $w_i \gets \frac{1}{m}\sum_{r=1}^m 
    \#\{g_j : \|g_i - g_j\| < \varepsilon_r\}$
\EndFor

\vspace{0.5em}
\State \textbf{/* Step 2: Weighted mean-shift with dynamic neighborhoods */}

\State Initialize shifted points $x_i^{(0)} \gets x_i$
\State Initialize cumulative displacement $D_i \gets 0$

\For{$t = 0$ to $T-1$}

    \State Build approximate $k$NN graph on $\{x_i^{(t)}\}$ using NNDescent

    \For{each point $x_i^{(t)}$}
        \State Compute weighted mean
        $\mu_i^{(t)} \gets 
        \sum_{j \in \mathcal{N}_k(x_i^{(t)})}
        \frac{w_j}{\sum_{\ell \in \mathcal{N}_k(x_i^{(t)})} w_\ell}
        x_j^{(t)}$

        \State Compute displacement
        $\delta_i^{(t)} \gets \|x_i^{(t)} - \mu_i^{(t)}\|$

        \State Update point
        $x_i^{(t+1)} \gets x_i^{(t)}
        + \eta \, \delta_i^{(t)}
        \frac{\mu_i^{(t)} - x_i^{(t)}}{\|\mu_i^{(t)} - x_i^{(t)}\| + \epsilon}$

        \State Accumulate $D_i \gets D_i + \delta_i^{(t)}$
    \EndFor

    \If{$\frac{1}{N}\sum_i \delta_i^{(t)} < \tau$}
        \State \textbf{break}
    \EndIf

\EndFor

\vspace{0.5em}
\State \textbf{/* Step 3: Anomaly scoring */}
\For{each $x_i$}
    \State $D_i \gets \sum_{t=0}^{T-1} \delta_i^{(t)}$
    \State $s_i \gets \sigma(\mathrm{scale}(D_i))$
\EndFor

\end{algorithmic}
\end{algorithm}

\subsection{Computational Complexity}\label{sec:complexity}

Let $N$ denote the number of samples, $d$ the feature dimension, $k$ the number of nearest neighbors, $T$ the maximum number of mean-shift iterations, $m$ the number of radii used for multi-scale density estimation, and $B$ the batch size used in the weight computation.

\paragraph{Neighborhood construction.}
During each mean-shift iteration, MSDE constructs a local neighborhood graph using approximate $k$-nearest neighbor search via NN-Descent. NN-Descent empirically scales close to linear time for fixed $k$ and typically incurs $\mathcal{O}(N k)$ time per construction. Since the neighborhood graph is recomputed at every iteration to reflect the evolving manifold structure, the total cost over $T$ iterations is $\mathcal{O}(T N k)$.

\paragraph{Fuzzy neighborhood graph and weight estimation.}
Empirical sample weights are computed prior to the mean-shift stage using a UMAP-based fuzzy neighborhood graph. The data are processed in mini-batches of size $B$. For each batch, approximate $k$NN search and fuzzy simplicial set construction incur $\mathcal{O}(B k)$ time. The subsequent multi-radius density estimation evaluates neighborhood counts at $m$ progressively decreasing radii using KD-tree range queries in the similarity space, resulting in an additional $\mathcal{O}(m B k)$ cost per batch. Aggregated across all batches, the total complexity of weight estimation is $\mathcal{O}(m N k)$.

\paragraph{Mean-shift updates.}
At each mean-shift iteration, a weighted neighborhood mean is computed for all $N$ samples using their $k$ nearest neighbors in the $d$-dimensional feature space. This results in a per-iteration cost of $\mathcal{O}(N k d)$. Over $T$ iterations, the total cost of the mean-shift updates is $\mathcal{O}(T N k d)$.

\paragraph{Overall complexity.}
Combining all components, the overall time complexity of the proposed MSDE implementation is
\[
\mathcal{O}\big(
T N k d + T N k + m N k
\big).
\]
For typical settings where $d \gg 1$, the dominant term is $\mathcal{O}(T N k d)$. In practice, MSDE uses small values of $T$ and $m$ (Table~\ref{tab:msde_parameters}), making it scalable to large tabular datasets.

Figure~\ref{fig:msde_pca_shift} in Appendix~\ref{secA2} illustrates the geometric behavior induced by MSDE on a representative dataset. As the neighborhood size $k$ increases, the mean-shift dynamics become smoother and more global, revealing stable manifold cores while amplifying the displacement of samples weakly supported by local density. Importantly, anomaly scoring in MSDE is not based on the final embedding, but on the cumulative displacement incurred during this density-driven evolution.

\section{Experimental Setup}\label{sec:exp_setup}

To rigorously evaluate the proposed Mean Shift Density Enhancement (MSDE) method, we adopt the experimental protocol of the \emph{ADBench} benchmark~\citep{jiang2022adbench}, which provides a large-scale, systematic comparison framework for anomaly detection algorithms. ADBench consists of forty six real-world tabular datasets spanning diverse application domains and anomaly characteristics, and evaluates $30$ anomaly detection methods under supervised, semi-supervised, and unsupervised settings. Since MSDE is a fully unsupervised method, our comparison focuses exclusively on unsupervised baselines.

\paragraph{Compared methods.}
We compare MSDE against $13$ representative unsupervised anomaly detection algorithms included in ADBench. These methods cover a broad spectrum of modeling assumptions, including distance-based detectors (KNN, LOF, COF), density-based methods (HBOS, COPOD, ECOD), clustering-based approaches (CBLOF, SOD), subspace and projection-based techniques (PCA, OCSVM), ensemble methods (Isolation Forest, LODA), and deep probabilistic models (DAGMM). This diverse selection ensures a fair and comprehensive assessment across classical, ensemble-based, and deep unsupervised paradigms. All methods are evaluated in a strictly unsupervised setting, with no access to anomaly labels during training or scoring. Following the ADBench protocol, experiments are conducted across all datasets using four realistic synthetic anomaly generation modes (global, local, cluster and dependency) across 3 random seeds as detailed in Table~\ref{tab:adbench_anomaly_types}.

\paragraph{Noise injection and robustness evaluation.}
In addition to varying anomaly types, we evaluate the robustness of all methods under different levels of feature noise, following the ADBench experimental design. Specifically, independent noise is injected into the data at six different noise ratios, ranging from noise-free settings to progressively higher noise levels. This perturbation simulates irrelevant feature corruption commonly encountered in real-world tabular datasets. Performance is reported separately for each noise level to assess the stability and degradation behavior of anomaly detectors under increasing noise.

\paragraph{Performance measures.}
We report results using three complementary evaluation metrics: Area Under the Receiver Operating Characteristic curve (AUC-ROC), Area Under the Precision–Recall curve (AUC-PR), and Precision@n.

\textbf{AUC-ROC} measures the ability of a detector to rank anomalous samples ahead of normal ones across all possible decision thresholds, making it a standard metric for anomaly detection benchmarks. However, AUC-ROC can be overly optimistic in highly imbalanced settings which are common in anomaly detection, although recent studies have reported contradictory findings regarding this effect~\citep{mcdermott_closer_2025}.

\textbf{AUC-PR} focuses on the precision–recall trade-off and is particularly sensitive to class imbalance. It emphasizes the correctness of anomaly rankings when anomalies constitute a small fraction of the data, thereby providing a more informative assessment of detector performance in practical scenarios.

\textbf{Precision@n} evaluates the fraction of true anomalies among the top-$n$ highest-scoring samples, where $n$ equals the number of ground-truth anomalies in each dataset. Unlike threshold-independent metrics, Precision@n directly measures the quality of the top-ranked detections and reflects real-world use cases in which only a limited number of flagged instances can be manually inspected. Although Precision@n is not originally reported in ADBench, we include it as a complementary metric to better assess extreme-tail anomaly ranking performance.

\paragraph{Implementation details.}
All baseline methods are implemented using the official ADBench codebase with default hyperparameters, ensuring consistency with prior evaluations. MSDE is implemented without dataset-specific tuning. Unless stated otherwise, all experiments are conducted using the default MSDE parameters described in Table~\ref{tab:msde_parameters}. This design choice emphasizes robustness and reproducibility across heterogeneous datasets, anomaly types and noise settings.

Our experimental setup enables a fair, comprehensive, and reproducible comparison of MSDE against established unsupervised anomaly detection methods under realistic and challenging benchmark conditions.

\begin{figure}[htbp!]
    \centering

    \begin{subfigure}{\textwidth}
        \centering
        \includegraphics[width=1\linewidth, height=4.8cm, keepaspectratio]{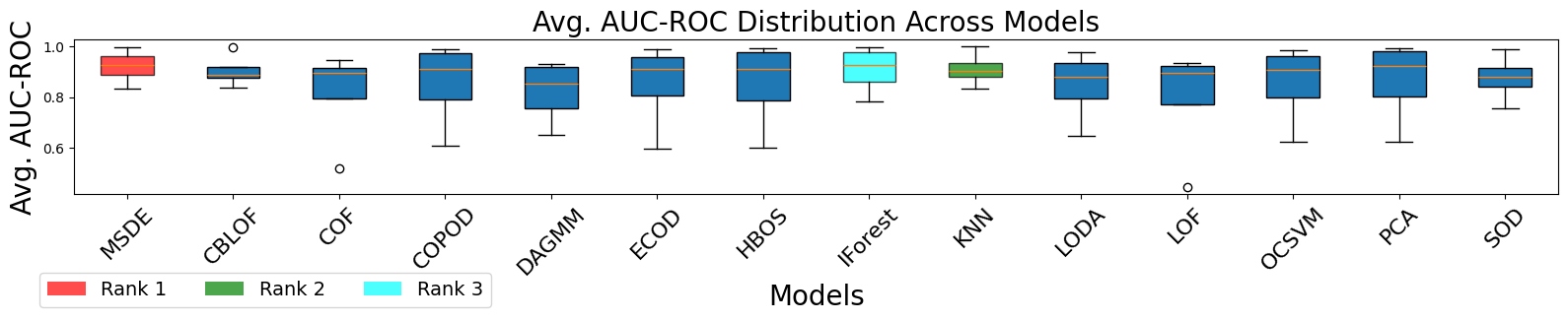}
        \caption{AUC-ROC}
        \label{fig:aucroc}
    \end{subfigure}

    \vspace{0.8em}

    \begin{subfigure}{\textwidth}
        \centering
        \includegraphics[width=1\linewidth, height=4.8cm, keepaspectratio]{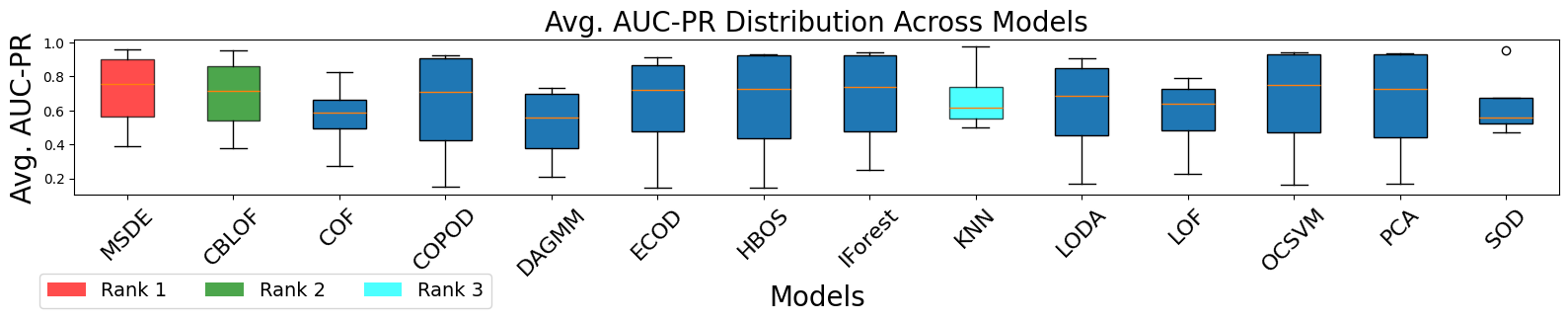}
        \caption{AUC-PR}
        \label{fig:aucpr}
    \end{subfigure}

    \vspace{0.8em}

    \begin{subfigure}{\textwidth}
        \centering
        \includegraphics[width=1\linewidth, height=4.8cm, keepaspectratio]{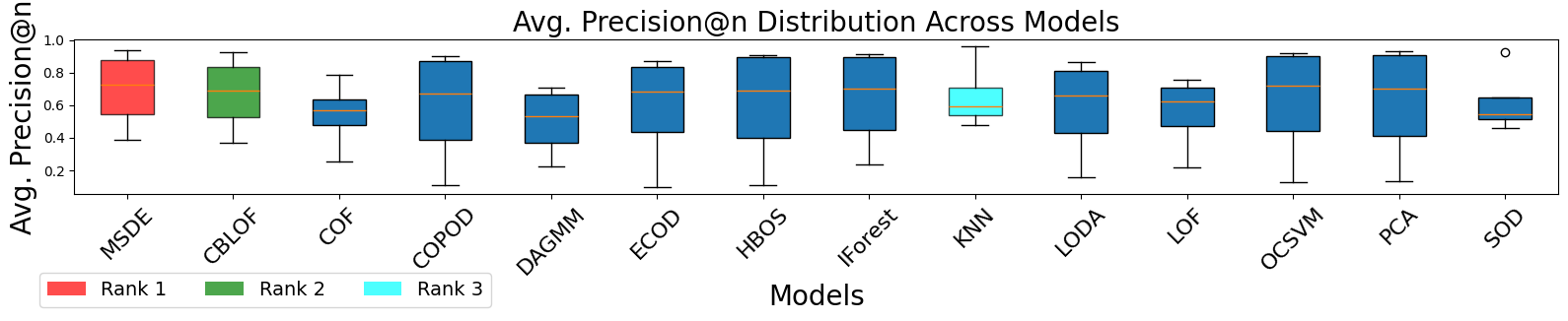}
        \caption{Precision@N}
        \label{fig:precision}
    \end{subfigure}

    \caption{Performance comparison of anomaly detection methods under no noise across evaluation metrics. Under the zero noise setting, MSDE outperforms other models on average over different anomaly types.}
    \label{fig:three_metrics}
\end{figure}

\section{Results}\label{sec:results}

We evaluate the proposed Mean Shift Density Enhancement (MSDE) method under the unsupervised ADBench protocol, considering both noise-free and noisy settings.
Results are averaged over all datasets and three random seeds, ensuring robust comparisons.

\paragraph{Performance across synthetic anomaly generation modes.}
To assess robustness across different anomaly mechanisms, we analyze MSDE’s performance separately under the four synthetic anomaly generation modes used in ADBench: dependency, cluster, global, and local anomalies. Figure~\ref{fig:three_metrics} shows our results, and Figure~\ref{fig:overall_results} presents the dataset wise AUC-ROC values under the zero noise setting. The full per-method results and top three ranks of the dataset-wise results are presented in Appendix~\ref{secA2} (Tables~\ref{tab:syntheticmode_multimodel} and \ref{tab:overall_results} respectively).

For \emph{global anomalies}, nearly all methods perform well due to the clear separation between anomalous and normal samples. MSDE ranks second across AUC-ROC, AUC-PR, and Precision@n, with performance (\(0.998/0.958/0.935\)) closely trailing KNN, which achieves the highest scores in this mode. 

In the \emph{local anomaly} setting, MSDE ranks fourth in AUC-ROC and third in both AUC-PR and Precision@n. Here, MSDE is outperformed primarily by LOF and KNN, which are explicitly designed to exploit local density ratios and nearest-neighbor distances. 

For \emph{cluster anomalies}, projection- and variance-based methods such as PCA, HBOS, OCSVM, and COPOD outperform MSDE by exploiting strong inter-cluster separability and global variance structure. These methods benefit from the presence of compact, well-separated anomalous clusters that violate global distributional assumptions. Although quite a strong performer in other modes, the performance of KNN is notably weak in this mode.

The most challenging setting is \emph{dependency anomalies}, where MSDE ranks between 5 and 6 across metrics. In this mode, MSDE is outperformed by methods such as COF, LOF, and KNN. Dependency anomalies violate feature correlations rather than marginal distributions, favoring detectors that encode relational structure more directly. 

While individual baselines excel at specific anomaly mechanisms, MSDE is the only method that consistently maintains stable performance across all four synthetic modes. MSDE avoids catastrophic performance degradation in any single mode and offers a stable, general-purpose solution for heterogeneous anomaly detection scenarios. Aggregating performance across all four synthetic modes (Figure~\ref{fig:three_metrics}), MSDE achieves the highest average AUC-ROC, AUC-PR, and Precision@n among all compared methods in the zero noise setting. Thus, MSDE offers balanced and consistently strong performance across diverse anomaly types.

Additionally, we 
extended the evaluation to include three top-performing supervised anomaly detection methods from the original ADBench benchmarking study. As expected, the best supervised model (CatBoost) achieved higher average performance; however, MSDE remained competitive 
in the more challenging unsupervised setting. We 
report the full numerical results from Figure~\ref{fig:three_metrics} along with the three supervised methods in Appendix~\ref{secA2} (Table~\ref{tab:syntheticmode_average}).

\paragraph{Statistical Significance.}
Table~\ref{tab:wilcoxon_all} presents the results of the one-sided Wilcoxon signed-rank test comparing MSDE against thirteen unsupervised anomaly detection baselines across the 46 benchmark datasets. Since all methods were evaluated under identical experimental settings, the Wilcoxon test provides an appropriate paired statistical comparison without assuming normality of performance differences. The results show that MSDE achieves statistically significant improvements ($p < 0.05$) over 12 out of the 13 compared methods in terms of AUC-ROC performance. In particular, large positive mean differences are observed against methods such as LOF, COF, DAGMM, and LODA, indicating substantial performance gains. Although MSDE also achieves a higher average AUC-ROC than KNN, 
this comes at a significantly higher $p$-value of $0.319$. Overall, these results demonstrate that the performance improvements achieved by MSDE are not only consistent but also statistically reliable across a wide range of datasets.

\begin{table}
\centering
\caption{Wilcoxon one-sided signed-rank test results comparing MSDE against thirteen
         unsupervised anomaly detection methods across 46 benchmark datasets for three
         evaluation metrics: AUC-ROC, AUC-PR, and Precision@$n$.
         Each row reports the mean score of the respective baseline, the mean score
         of MSDE, the signed mean difference $\Delta = \bar{s}_{\text{MSDE}} - \bar{s}_{\text{baseline}}$,
         the Wilcoxon statistic~$W$, and the one-sided $p$-value. Results are considered significant if $p < 0.05$.}
\label{tab:wilcoxon_all}
\begin{tabular}{llccccc}
\toprule
\textbf{Metric} & \textbf{Baseline} & \textbf{Score} & \textbf{MSDE} & \textbf{$\Delta$} & \textbf{$W$} & \textbf{$p$-value} \\
\midrule

\multirow{13}{*}{\rotatebox[origin=c]{90}{\textbf{AUC-ROC}}}
 & DAGMM   & 0.829 & 0.925 & $+$0.096 & 125578.0 & $< 10^{-60}$$\,$ \\
 & LODA    & 0.855 & 0.925 & $+$0.070 & 100808.0 & $1.20 \times 10^{-50}$$\,$ \\
 & ECOD    & 0.863 & 0.925 & $+$0.062 &  85906.0 & $4.61 \times 10^{-41}$$\,$ \\
 & HBOS    & 0.864 & 0.925 & $+$0.062 &  77171.0 & $2.22 \times 10^{-37}$$\,$ \\
 & COPOD   & 0.866 & 0.925 & $+$0.060 &  76598.0 & $7.80 \times 10^{-37}$$\,$ \\
 & OCSVM   & 0.866 & 0.925 & $+$0.059 &  74408.5 & $2.90 \times 10^{-36}$$\,$ \\
 & PCA     & 0.875 & 0.925 & $+$0.050 &  75294.5 & $4.61 \times 10^{-32}$$\,$ \\
 & IForest & 0.914 & 0.925 & $+$0.012 &  67396.5 & $1.21 \times 10^{-18}$$\,$ \\
 & COF     & 0.812 & 0.925 & $+$0.114 &  85553.0 & $2.01 \times 10^{-17}$$\,$ \\
 & SOD     & 0.877 & 0.925 & $+$0.049 &  79589.0 & $3.83 \times 10^{-17}$$\,$ \\
 & LOF     & 0.791 & 0.925 & $+$0.134 &  71494.0 & $6.23 \times 10^{-10}$$\,$ \\
 & CBLOF   & 0.906 & 0.924 & $+$0.019 &  54767.5 & $7.02 \times 10^{-8}$$\,$ \\
\cmidrule(l){2-7}
 & KNN     & 0.911 & 0.925 & $+$0.014 &  53420.0 & $0.319$ \\
\midrule

\multirow{13}{*}{\rotatebox[origin=c]{90}{\textbf{AUC-PR}}}
 & DAGMM   & 0.529 & 0.727 & $+$0.198 & 128408.5 & $< 10^{-60}$$\,$ \\
 & LODA    & 0.631 & 0.727 & $+$0.096 &  98640.0 & $1.65 \times 10^{-44}$$\,$ \\
 & COPOD   & 0.644 & 0.727 & $+$0.083 &  75362.0 & $1.66 \times 10^{-33}$$\,$ \\
 & ECOD    & 0.645 & 0.727 & $+$0.082 &  81366.0 & $6.67 \times 10^{-32}$$\,$ \\
 & PCA     & 0.661 & 0.727 & $+$0.066 &  75120.0 & $1.03 \times 10^{-31}$$\,$ \\
 & HBOS    & 0.652 & 0.727 & $+$0.075 &  76938.0 & $4.40 \times 10^{-29}$$\,$ \\
 & OCSVM   & 0.670 & 0.727 & $+$0.057 &  70172.0 & $1.73 \times 10^{-27}$$\,$ \\
 & IForest & 0.684 & 0.727 & $+$0.043 &  72964.5 & $3.22 \times 10^{-22}$$\,$ \\
 & COF     & 0.568 & 0.727 & $+$0.159 &  82657.0 & $3.71 \times 10^{-14}$$\,$ \\
 & SOD     & 0.641 & 0.727 & $+$0.086 &  74793.0 & $2.03 \times 10^{-11}$$\,$ \\
 & LOF     & 0.574 & 0.727 & $+$0.153 &  69458.0 & $3.94 \times 10^{-8}$$\,$ \\
 & CBLOF   & 0.700 & 0.724 & $+$0.023 &  53375.0 & $2.30 \times 10^{-6}$$\,$ \\
\cmidrule(l){2-7}
 & KNN     & 0.681 & 0.727 & $+$0.046 &  52149.0 & $0.493$ \\
\midrule

\multirow{13}{*}{\rotatebox[origin=c]{90}{\textbf{Precision@$n$}}}
 & DAGMM   & 0.511 & 0.707 & $+$0.197 & 108917.5 & $3.75 \times 10^{-60}$$\,$ \\
 & LODA    & 0.601 & 0.707 & $+$0.107 &  81930.0 & $7.88 \times 10^{-40}$$\,$ \\
 & COPOD   & 0.608 & 0.707 & $+$0.099 &  64005.5 & $2.72 \times 10^{-34}$$\,$ \\
 & ECOD    & 0.605 & 0.707 & $+$0.102 &  67153.5 & $3.07 \times 10^{-33}$$\,$ \\
 & HBOS    & 0.621 & 0.707 & $+$0.087 &  58372.5 & $1.32 \times 10^{-26}$$\,$ \\
 & PCA     & 0.637 & 0.707 & $+$0.071 &  55597.0 & $3.59 \times 10^{-24}$$\,$ \\
 & OCSVM   & 0.642 & 0.707 & $+$0.066 &  50581.0 & $6.66 \times 10^{-22}$$\,$ \\
 & IForest & 0.655 & 0.707 & $+$0.052 &  51962.5 & $5.63 \times 10^{-19}$$\,$ \\
 & COF     & 0.545 & 0.707 & $+$0.163 &  74122.0 & $7.71 \times 10^{-17}$$\,$ \\
 & SOD     & 0.621 & 0.707 & $+$0.086 &  62693.5 & $6.36 \times 10^{-12}$$\,$ \\
 & LOF     & 0.555 & 0.707 & $+$0.152 &  63652.0 & $5.83 \times 10^{-10}$$\,$ \\
 & CBLOF   & 0.679 & 0.704 & $+$0.026 &  36030.5 & $1.22 \times 10^{-5}$$\,$ \\
\cmidrule(l){2-7}
 & KNN     & 0.659 & 0.707 & $+$0.049 &  44474.0 & $0.091$ \\
\bottomrule
\end{tabular}
\end{table}

\paragraph{Performance under noise-free conditions and robustness to increasing noise levels.}

\begin{figure}[t]
    \centering
    \includegraphics[width=\textwidth]{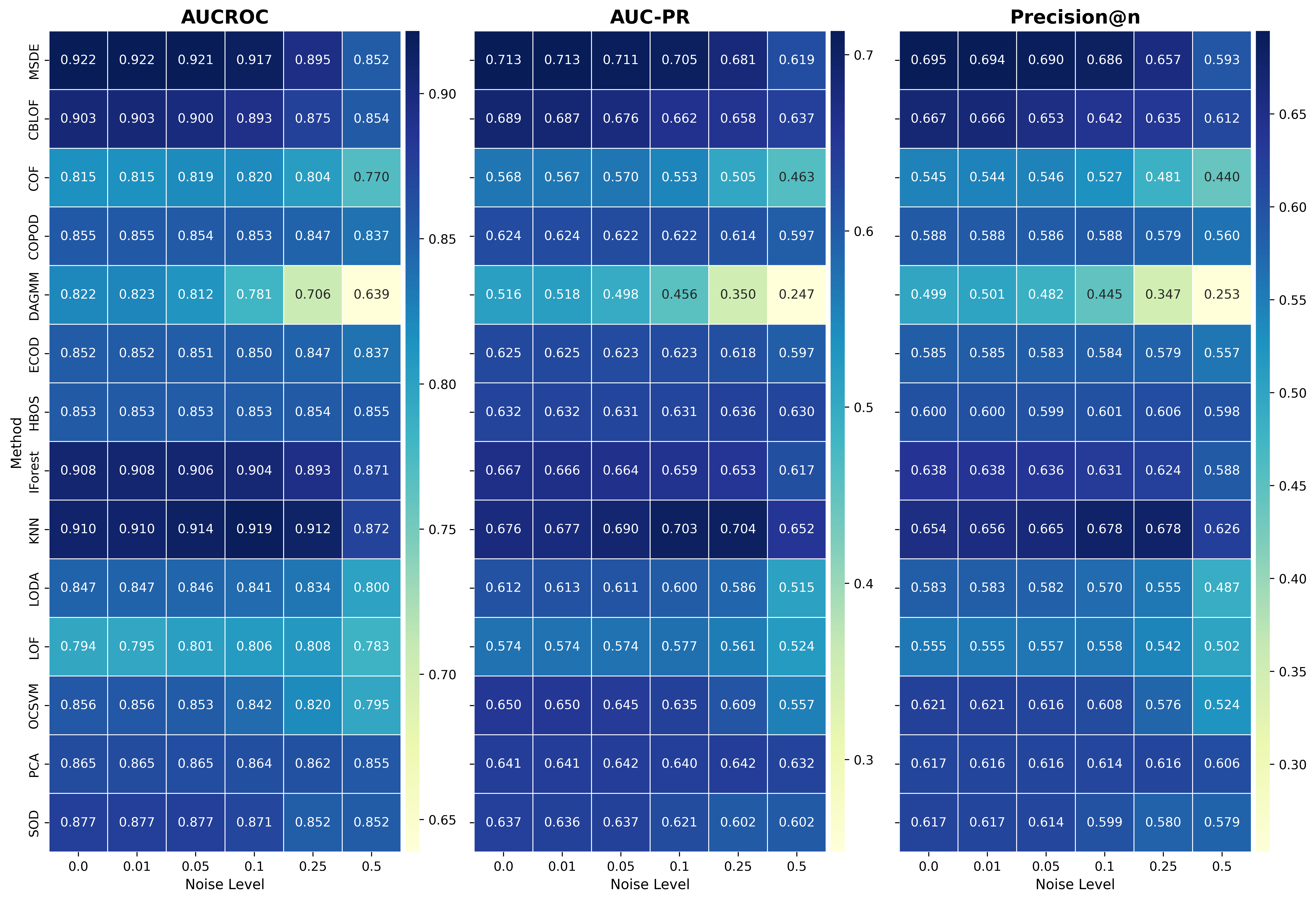}
    \caption{Performance heatmaps of all compared methods under varying noise levels (0.00, 0.01, 0.05, 0.10, 0.25, 0.50) averaged over the four different synthetic modes for AUCROC, AUC-PR, and Precision@n metrics.}
    \label{fig:noisePerformance}
\end{figure}

Figure~\ref{fig:noisePerformance} illustrates the performance trends of all compared anomaly detection methods under progressively increasing feature noise levels ($0.00$, $0.01$, $0.05$, $0.10$, $0.25$, $0.50$), averaged across the four synthetic anomaly generation modes. The heatmaps provide a direct visual comparison of robustness across AUC-ROC, AUC-PR, and Precision@n metrics.

Overall, all methods experience performance degradation as the noise ratio increases. However, MSDE demonstrates comparatively stable behavior and maintains strong performance across all noise settings. At low to moderate noise levels ($0.00$--$0.10$), MSDE consistently achieves among the highest scores across all three evaluation metrics, particularly maintaining AUC-ROC values above $0.91$, indicating strong resilience to mild feature perturbations.

As the corruption level increases further ($0.25$ and $0.50$), the performance gap between methods narrows. In these highly noisy scenarios, neighborhood-based approaches such as KNN and clustering-based methods such as CBLOF occasionally achieve competitive or superior AUC-PR and Precision@n values. Nevertheless, MSDE remains consistently competitive and avoids the severe degradation observed in reconstruction-based methods such as DAGMM and local density estimators such as LOF and COF.

The heatmaps further reveal that several traditional methods exhibit substantial sensitivity to increasing noise, reflected by rapidly fading performance values at higher corruption levels. In contrast, MSDE shows a more gradual decline across all metrics, highlighting its robustness and stability under noisy data conditions.

\paragraph{Ablation study.}
We also performed an ablation study to isolate the contributions of the fuzzy simplicial graph construction and the multi-radius weighting module. Specifically, we ablate the fuzzy simplicial nearest-neighbour graph by replacing it with a symmetric, unweighted k-nearest-neighbour graph, while keeping the neighbourhood size fixed. Separately, we ablate the multi-radius weighting scheme by using a single radius weight. This radius is computed via the same binary search and satisfiability criterion as in the full model; however, neighbourhood counts are computed only once, and no radius decay or aggregation is applied. These two ablations are evaluated independently, and model performance is reported for each setting. In addition, we report the performance of MSDE (opt), in which hyperparameters are tuned independently for each synthetic anomaly mode using Optuna, with 10 trials per mode.

\begin{table}[htbp!]
\centering
\caption{Performance comparison of MSDE and its ablated variants under different synthetic anomaly modes. The table reports results for the main MSDE model, an optimized configuration, a single-radius weight estimation variant, and a simple binary graph variant, highlighting the impact of graph construction and multi-scale density estimation.}
\label{tab:ablation_cases}
\begin{tabular}{llccc}

\toprule
Method & Synthetic Modes & AUC-ROC $\uparrow$ & AUC-PR $\uparrow$ & Precision@n $\uparrow$ \\
\midrule
MSDE & \multirow{4}{*}{Dependency} & 0.834 ± 0.140 & 0.390 ± 0.280 & 0.389 ± 0.296 \\
MSDE (opt) & & \textbf{0.880 ± 0.124} & \textbf{0.521 ± 0.308} & \textbf{0.515 ± 0.300} \\
MSDE (simple graph) & & 0.842 ± 0.133 & 0.383 ± 0.278 & 0.383 ± 0.293 \\
MSDE (single radius) & & 0.834 ± 0.139 & 0.391 ± 0.282 & 0.390 ± 0.296 \\

\midrule
MSDE & \multirow{4}{*}{Cluster} & \textbf{0.952 ± 0.094} & \textbf{0.886 ± 0.190} & \textbf{0.861 ± 0.202} \\
MSDE (opt)& & 0.944 ± 0.101 & \textbf{0.868 ± 0.201} & 0.841 ± 0.211   \\
MSDE (simple graph) & & 0.946 ± 0.099 & 0.884 ± 0.191 & 0.852 ± 0.211 \\
MSDE (single radius) & & 0.948 ± 0.102 & 0.879 ± 0.200 & 0.855 ± 0.213 \\

\midrule
MSDE & \multirow{4}{*}{Global} & \textbf{0.998 ± 0.005} & \textbf{0.958 ± 0.132} & 0.935 ± 0.143 \\
MSDE (opt) & & \textbf{0.998 ± 0.004} & \textbf{0.958 ± 0.133} & \textbf{0.937 ± 0.143} \\
MSDE (simple graph) & & 0.997 ± 0.005 & 0.951 ± 0.142 & 0.926 ± 0.155 \\
MSDE (single radius) & & 0.997 ± 0.005 & 0.956 ± 0.134 & 0.930 ± 0.151 \\

\midrule
MSDE & \multirow{4}{*}{Local} & 0.904 ± 0.096 & 0.620 ± 0.318 & 0.593 ± 0.300 \\
MSDE (opt) & & \textbf{0.912 ± 0.092} & \textbf{0.635 ± 0.312} & \textbf{0.608 ± 0.298}  \\
MSDE (simple graph) & & 0.898 ± 0.098 & 0.593 ± 0.319 & 0.560 ± 0.310 \\
MSDE (single radius) & & 0.904 ± 0.096 & 0.620 ± 0.318 & 0.591 ± 0.301 \\
\bottomrule

\end{tabular}
\end{table}

Results are reported in Table~\ref{tab:ablation_cases}. We find that the fuzzy simplicial graph yields higher AUC-ROC across all anomaly types except dependency anomalies. However, the improvement in performance of the simple graph on dependency anomalies is not reflected in the AUC-PR or Precision@n metrics. This indicates that the fuzzy graph particularly benefits detection when anomalies align with the dependency structure of the underlying data. Additionally, the full model consistently outperforms the single-radius weighting variant across all anomaly types and evaluation metrics, with the exception of dependency anomalies, where the two approaches achieve comparable performance. These results support the effectiveness of the proposed multi-radius weighting scheme.

\begin{figure}[htbp!]
\centering
\includegraphics[width=0.95\linewidth]{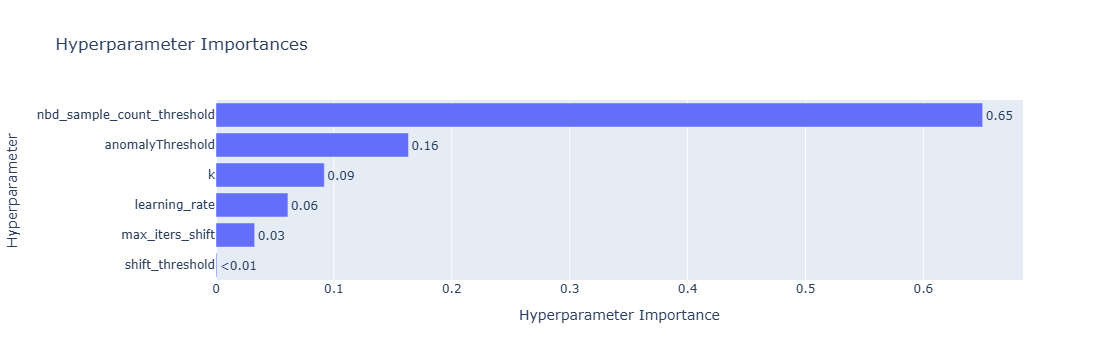}
\caption{Optuna-based hyperparameter importance analysis for MSDE on the Credit Card dataset.}
\label{fig:hyperparam_importance}
\end{figure}

We also peform a sensitivity analyis (details in Appendix~\ref{app:hyperparam_importance}) by estimating the importance of each hyperparameter on the objective function using functional analysis of variance, or fANOVA~\citep{fanova}, on the Credit Fraud Detection dataset~\citep{creditFraud}. The corresponding normalized importance values are presented in Figure~\ref{fig:hyperparam_importance}.

\section{Discussion}\label{sec:discussion}

\begin{figure}[htbp!]
    \centering
    \begin{minipage}{0.45\textwidth}
        \includegraphics[width=\linewidth]{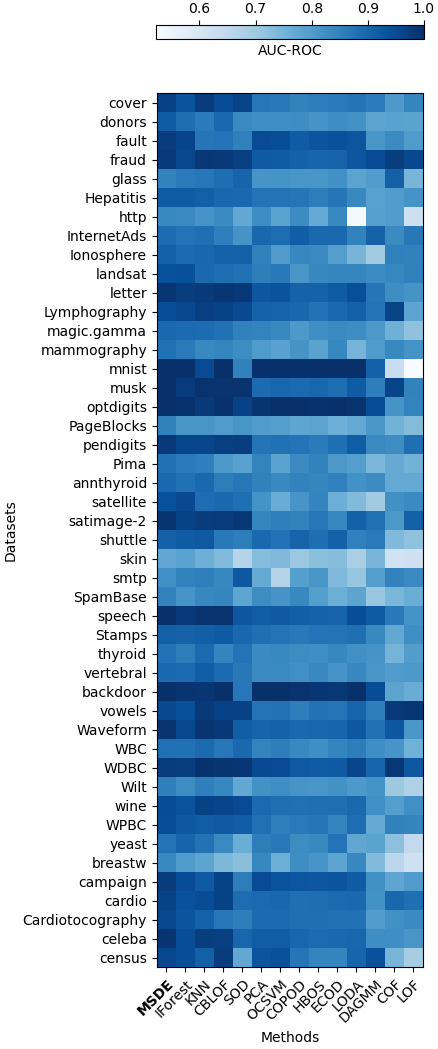}
        \caption{Dataset wise average AUC-ROC values for anomaly detection methods across all four synthetic anomaly generation modes under the zero noise setting}
        \label{fig:overall_results}
    \end{minipage}\hfill
    \begin{minipage}{0.53\textwidth}
        \includegraphics[width=\linewidth]{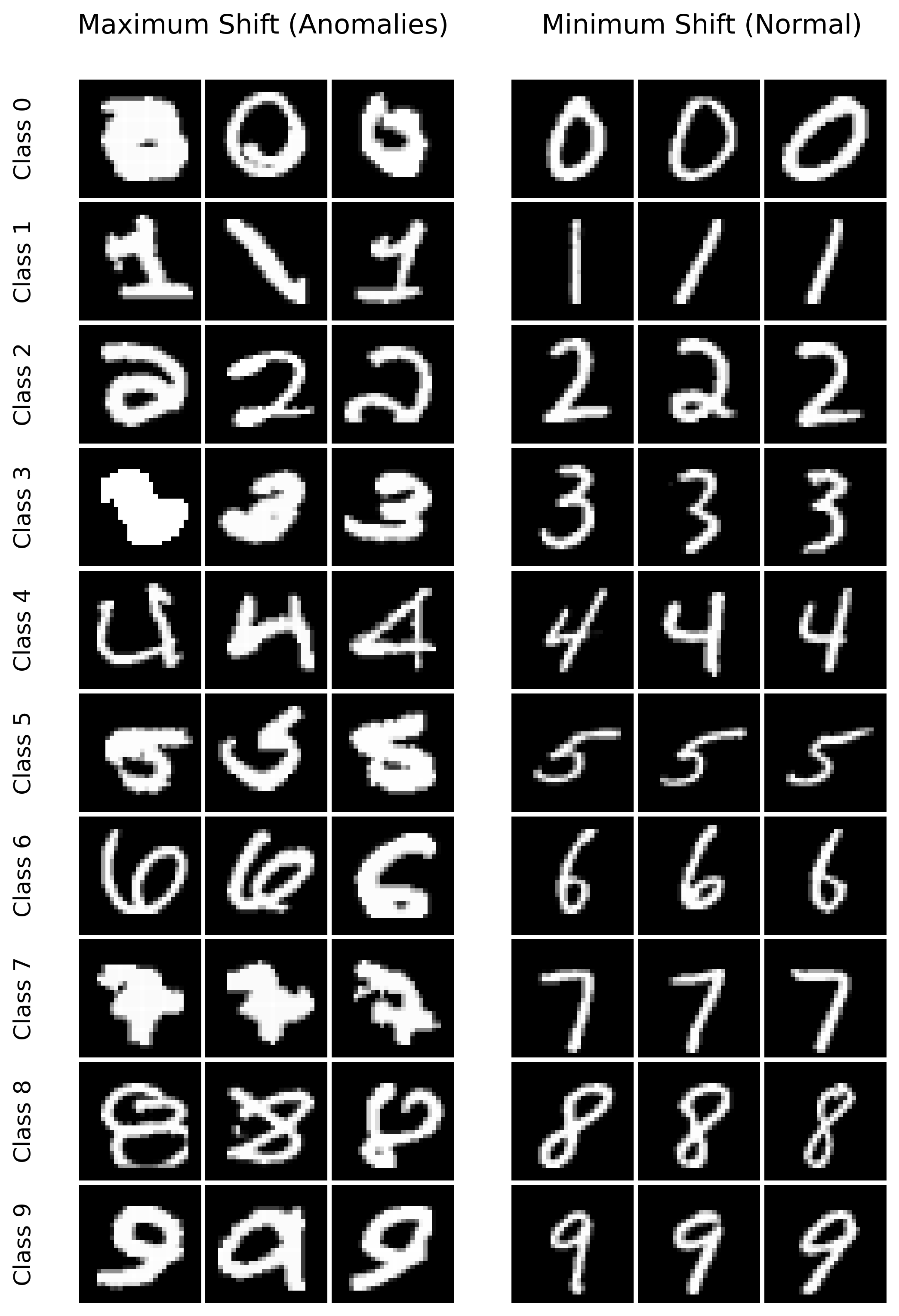}
        \caption{
        Qualitative illustration of MSDE anomaly detection on the MNIST dataset.
        \textbf{Left:} samples with the largest cumulative MSDE displacement (high anomaly scores).
        \textbf{Right:} samples with the smallest displacement (low anomaly scores).
        Across all digit classes, high-shift samples correspond to poorly written, ambiguous, or structurally irregular digits, while low-shift samples represent clean, prototypical instances.
        This demonstrates that MSDE captures semantic irregularity through geometric instability, despite operating in a fully unsupervised manner.
        }
        \label{fig:msde_mnist_shift}
    \end{minipage}
\end{figure}

\paragraph{Effectiveness and Novelty of the MSDE Approach for Anomaly Detection.}
The experimental results indicate that MSDE performs consistently well across all four synthetic anomaly generation modes and remains competitive across evaluation metrics and noise regimes. In particular, MSDE achieves high average performance across dependency, cluster, global, and local anomaly settings, and maintains strong performance under noise levels up to 0.1 (Figure~\ref{fig:noisePerformance}). Additionally, we performed a small scalability analysis of MSDE on the Credit Card Fraud Detection dataset~\citep{creditFraud} ($N=284,807$) and found that MSDE is an order of magnitude faster than computationally intensive methods like KNN, DAGMM, SOD, and OCSVM, while maintaining its competitiveness in terms of AUC-ROC (Figure~\ref{fig:runtime_creditcard}). Full details can be found in Appendix~\ref{app:scalability}.

\begin{figure}[t]
    \centering
    \includegraphics[width=\textwidth]{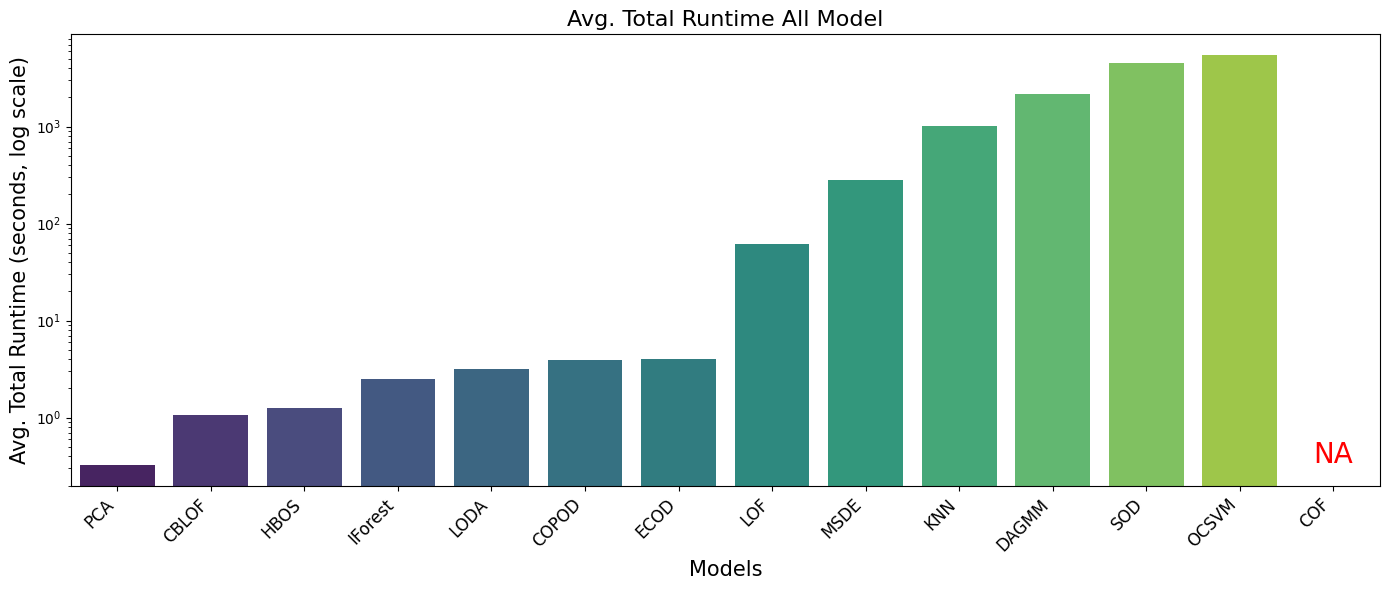}
    \caption{Average total runtime (log scale) of anomaly detection models on the Credit Card Fraud dataset. COF failed due to excessive memory requirements and is therefore excluded.}
    \label{fig:runtime_creditcard}
\end{figure}

MSDE scores anomalies based on cumulative geometric displacement induced by weighted mean-shift iterations rather than static distance or reconstruction error. Normal samples converge rapidly to nearby density modes and exhibit small cumulative movement, whereas anomalous samples undergo larger shifts before stabilization. This displacement-based scoring is a novel approach in the domain of unsupervised anomaly detection.

Figure~\ref{fig:msde_mnist_shift} provides an intuitive interpretation of MSDE anomaly scores on image data. Digits that are visually ambiguous, distorted, or atypically written undergo significantly larger cumulative shifts, whereas clean and canonical digit instances remain stable. This behavior aligns with the core hypothesis of MSDE: normal samples are supported by strong local density and therefore resist displacement, while anomalous samples exhibit pronounced geometric instability.

\paragraph{Limitations.}
Despite its strengths, MSDE has several limitations. First, the method relies on nearest-neighbor graph construction and repeated mean-shift updates, which introduce computational overhead compared to simple histogram- or projection-based detectors. Although the use of approximate kNN search and a small number of shift iterations keeps the method practical for medium- to large-scale tabular datasets, further optimization would be required for extremely large or streaming data scenarios.

Second, MSDE assumes that meaningful local neighborhood structure exists in the feature space. In datasets where features are weakly informative or dominated by noise, the constructed manifold may be unreliable, limiting the effectiveness of density enhancement. 
We also observed that for very high noise levels (0.25 and 0.5) performance of MSDE degrades compared to 
lower noise levels (for further details see Table~\ref{tab:noise_multimodel} in Appendix~\ref{secA2}). This limitation is shared by many geometry-based anomaly detection methods.

\paragraph{Beyond anomaly detection: broader applications of density enhancement.}

Beyond anomaly detection, MSDE introduces a general paradigm for density-driven geometric evolution of data. Sample trajectories under mean shift encode rich information about the underlying data manifold, rather than serving solely as an intermediate step for anomaly scoring.

The paths traced as points move toward local density modes can be viewed as manifold flow lines, revealing shared local structure, boundary regions, and transitional states. This trajectory-level information can be exploited directly for clustering in high-dimensional feature space, without explicit dimensionality reduction.

The direction and magnitude of feature-space shifts further provide a principled measure of feature relevance: dimensions inducing large displacements correspond to discriminative or unstable features, while stable coordinates remain largely unchanged. As shown in Appendix~\ref{secA2} Figure~\ref{fig:mnist_feature_shift}, MSDE concentrates displacement on semantically meaningful stroke regions in MNIST, supporting its interpretation as a structure-aware density enhancement process. The resulting density-regularized representations suppress noise and can benefit downstream tasks such as clustering and classification.

Overall, MSDE defines a structure-aware data transformation that reveals latent geometric organization without modifying the ambient feature space, enabling integration with representation learning and graph-based models.

\color{black}

\section{Conclusion}\label{sec:conclusion}
In this work, we introduced Mean Shift Density Enhancement, a non-parametric, fully unsupervised anomaly detection method that 
employs adaptive neighborhood modeling and iterative geometric refinement. Extensive evaluation on the ADBench benchmark demonstrates that MSDE achieves competitive or superior performance compared to established unsupervised baselines across multiple anomaly types, evaluation metrics, and noise levels. By framing anomaly detection as a density-driven dynamic process on the data manifold, MSDE provides both strong empirical performance and an interpretable geometric intuition. These properties make MSDE a promising and versatile tool for unsupervised anomaly detection, motivating further exploration of density enhancement techniques in broader machine learning applications.

\backmatter

\bmhead{Supplementary information}

All source code required to reproduce the experiments presented in this paper is included in the supplementary material as a ZIP archive. The numerical values for Figure~\ref{fig:overall_results} along with standard deviations that complete Table~\ref{tab:overall_results} are provided in the file ``\texttt{overall\_aucroc\_rankings.csv}". A cover letter summarising the main claims and results of the paper is also provided.




\section*{Statements and Declarations}

\subsection*{Funding} 
O.W. acknowledges support from the German Research Foundation (DFG) FK515800538 (learning convex data spaces).
\subsection*{Competing interests} 
The authors have no competing interests to declare that are relevant to the content of this article.


\bibliography{sn-bibliography}

\begin{appendices}

\section{Additional detailed results}\label{secA2}%

\begin{figure}[H]
    \centering

    \begin{subfigure}[t]{0.48\textwidth}
        \centering
        \includegraphics[width=\linewidth]{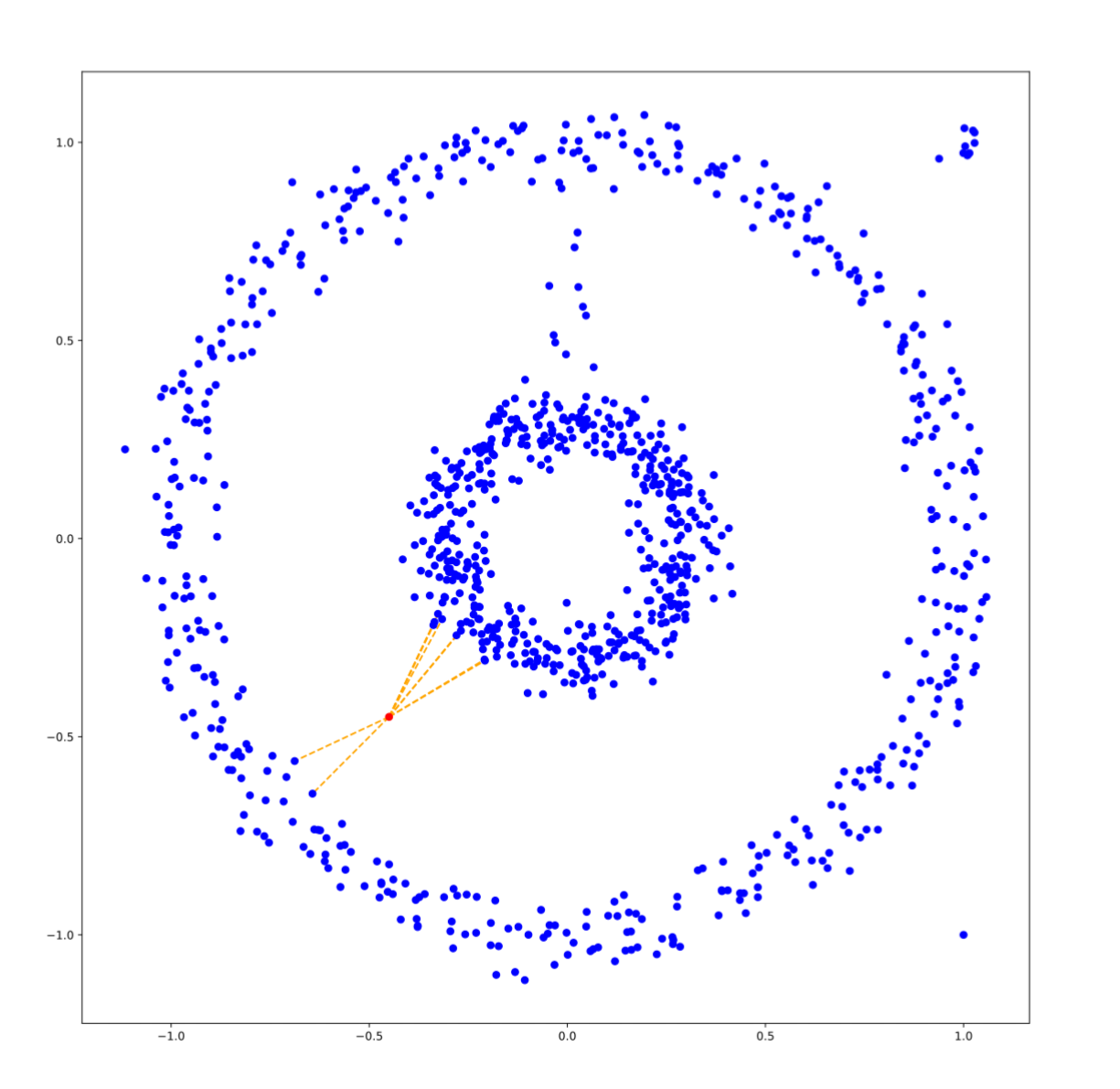}
        \caption{\textbf{Initial data representation.} 
        Input samples embedded in the original feature space prior to manifold-aware processing.}
        \label{fig:msde_step1}
    \end{subfigure}
    \hfill
    \begin{subfigure}[t]{0.48\textwidth}
        \centering
        \includegraphics[width=\linewidth]{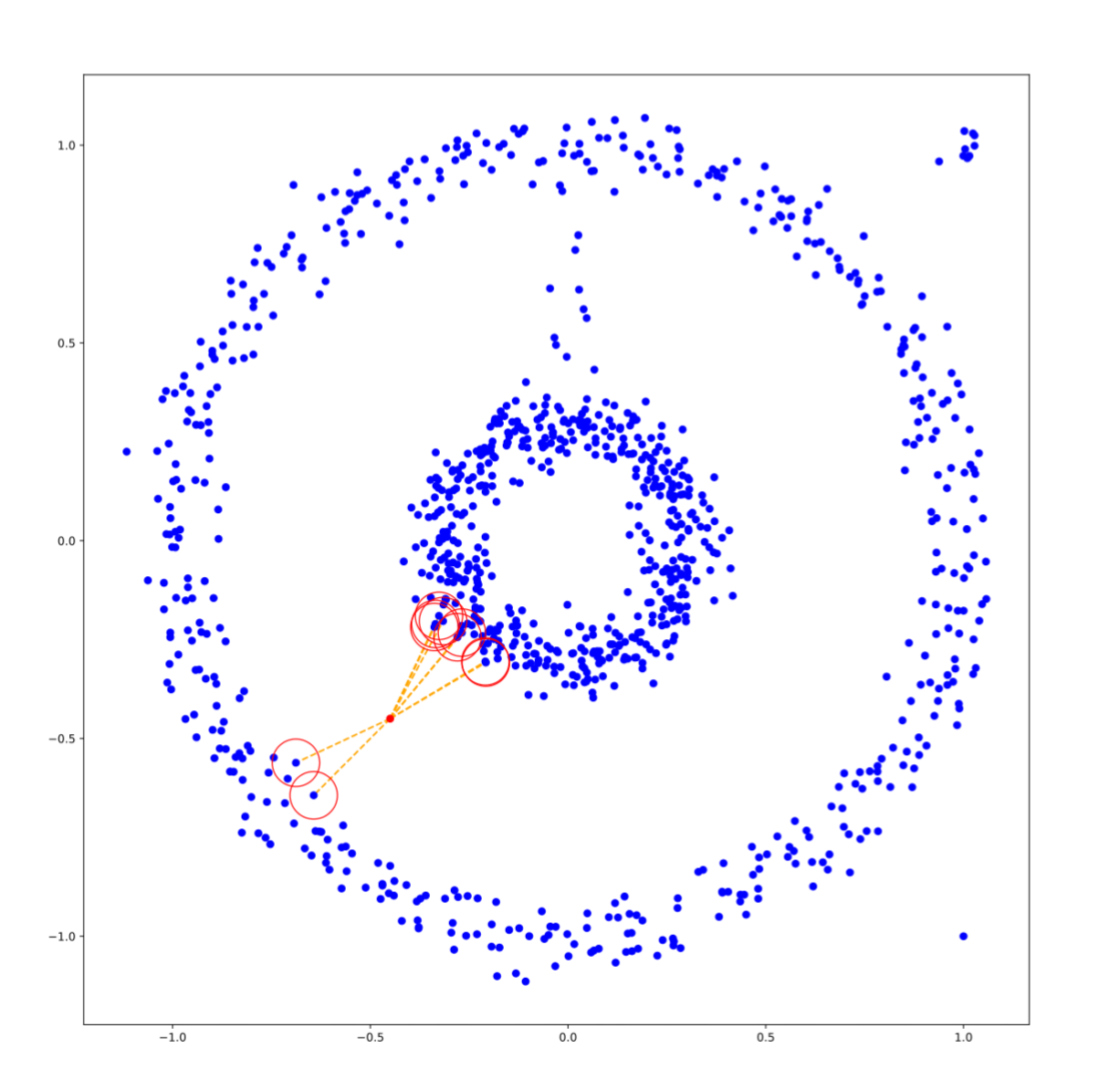}
        \caption{\textbf{Manifold similarity construction.} 
        A fuzzy neighborhood graph captures local geometric relationships among samples.}
        \label{fig:msde_step2}
    \end{subfigure}

    \vspace{0.5em}

    \begin{subfigure}[t]{0.48\textwidth}
        \centering
        \includegraphics[width=\linewidth]{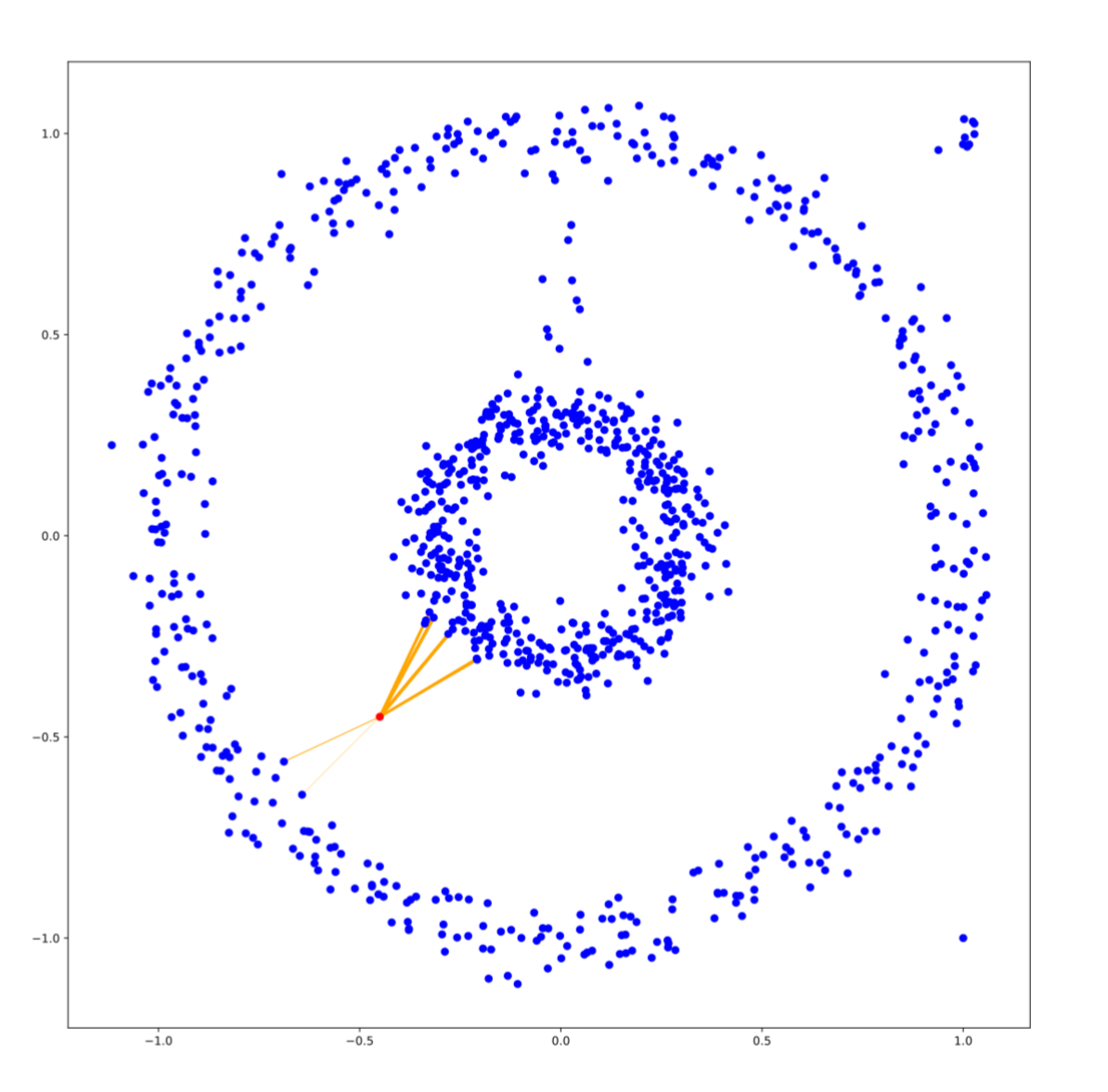}
        \caption{\textbf{Weighted mean-shift evolution.} 
        Samples are iteratively shifted toward dense manifold regions using empirical neighborhood weights.}
        \label{fig:msde_step3}
    \end{subfigure}
    \hfill
    \begin{subfigure}[t]{0.48\textwidth}
        \centering
        \includegraphics[width=\linewidth]{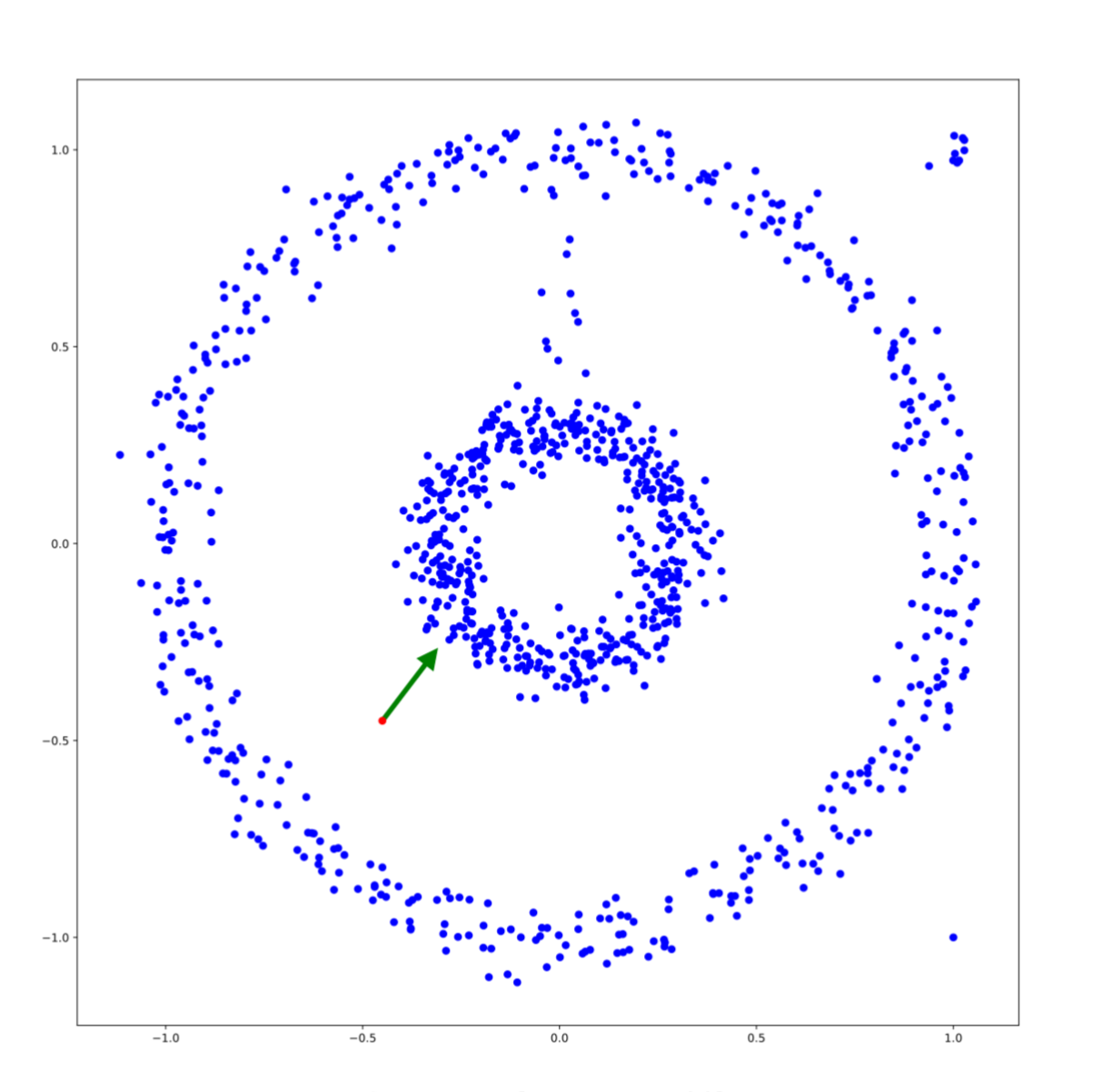}
        \caption{\textbf{Anomaly scoring via cumulative displacement.} 
        Points exhibiting larger total movement are assigned higher anomaly scores.}
        \label{fig:msde_step4}
    \end{subfigure}

    \caption{\textbf{Step-by-step illustration of the proposed MSDE algorithm.} 
    The method constructs a manifold-aware similarity structure, performs weighted mean-shift updates, and quantifies anomaly likelihood using cumulative point displacement.}
    \label{fig:msde_steps}
\end{figure}

\subsection{Algorithm Overview}

Figure~\ref{fig:msde_steps} illustrates the key stages of the proposed MSDE framework. The algorithm begins by representing the input data in its original feature space (Fig.~\ref{fig:msde_step1}). A manifold-aware similarity graph is then constructed using fuzzy neighborhood relationships, capturing local geometric structure (Fig.~\ref{fig:msde_step2}).

Next, an iterative weighted mean-shift procedure is applied, where each point is updated based on the empirical density of its neighborhood on the learned manifold (Fig.~\ref{fig:msde_step3}). This process causes samples in dense regions to stabilize quickly, while isolated samples undergo larger displacements.

Finally, the cumulative displacement across iterations is used as an anomaly score
(Fig.~\ref{fig:msde_step4}), with larger movements indicating higher anomaly likelihood.

\begin{table}[htbp!]
\centering
\caption{Average performance of anomaly detection methods across all four synthetic anomaly generation modes under zero noise setting.}
\label{tab:syntheticmode_average}
\begin{tabular}{lccc}
\toprule
Method & Avg AUC-ROC $\uparrow$ & Avg AUC-PR $\uparrow$ & Avg Precision@n $\uparrow$ \\
\midrule
MSDE   & \textbf{0.922 ± 0.070} & \textbf{0.714 ± 0.260} & \textbf{0.695 ± 0.251} \\
KNN    & 0.910 ± 0.068 & 0.676 ± 0.210 & 0.655 ± 0.213 \\
IForest& 0.908 ± 0.097 & 0.667 ± 0.329 & 0.638 ± 0.323 \\
CBLOF  & 0.904 ± 0.067 & 0.689 ± 0.255 & 0.667 ± 0.246 \\
SOD    & 0.878 ± 0.095 & 0.637 ± 0.215 & 0.617 ± 0.208 \\
PCA    & 0.865 ± 0.170 & 0.641 ± 0.368 & 0.617 ± 0.375 \\
OCSVM  & 0.856 ± 0.164 & 0.650 ± 0.367 & 0.621 ± 0.371 \\
COPOD  & 0.855 ± 0.176 & 0.624 ± 0.369 & 0.588 ± 0.373 \\
HBOS   & 0.853 ± 0.180 & 0.632 ± 0.373 & 0.600 ± 0.378 \\
ECOD   & 0.852 ± 0.177 & 0.625 ± 0.349 & 0.585 ± 0.354 \\
LODA   & 0.847 ± 0.143 & 0.612 ± 0.332 & 0.583 ± 0.321 \\
DAGMM  & 0.822 ± 0.129 & 0.516 ± 0.242 & 0.499 ± 0.221 \\
COF    & 0.815 ± 0.197 & 0.568 ± 0.228 & 0.545 ± 0.220 \\
LOF    & 0.794 ± 0.233 & 0.574 ± 0.248 & 0.555 ± 0.240 \\
\bottomrule
Sup. Method & Avg AUC-ROC $\uparrow$ & Avg AUC-PR $\uparrow$ & Avg Precision@n $\uparrow$ \\
\midrule
CatB   & \textbf{0.949 ± 0.099} & \textbf{0.806 ± 0.282} & \textbf{0.778 ± 0.278}\\
XGB    & 0.866 ± 0.179 & 0.672 ± 0.363 & 0.644 ± 0.348 \\
MLP    & 0.688 ± 0.291 & 0.516 ± 0.381 & 0.491 ± 0.377 \\
\bottomrule
\end{tabular}
\end{table}

\begin{figure}[htbp!]
    \centering
    \includegraphics[width=0.95\linewidth]{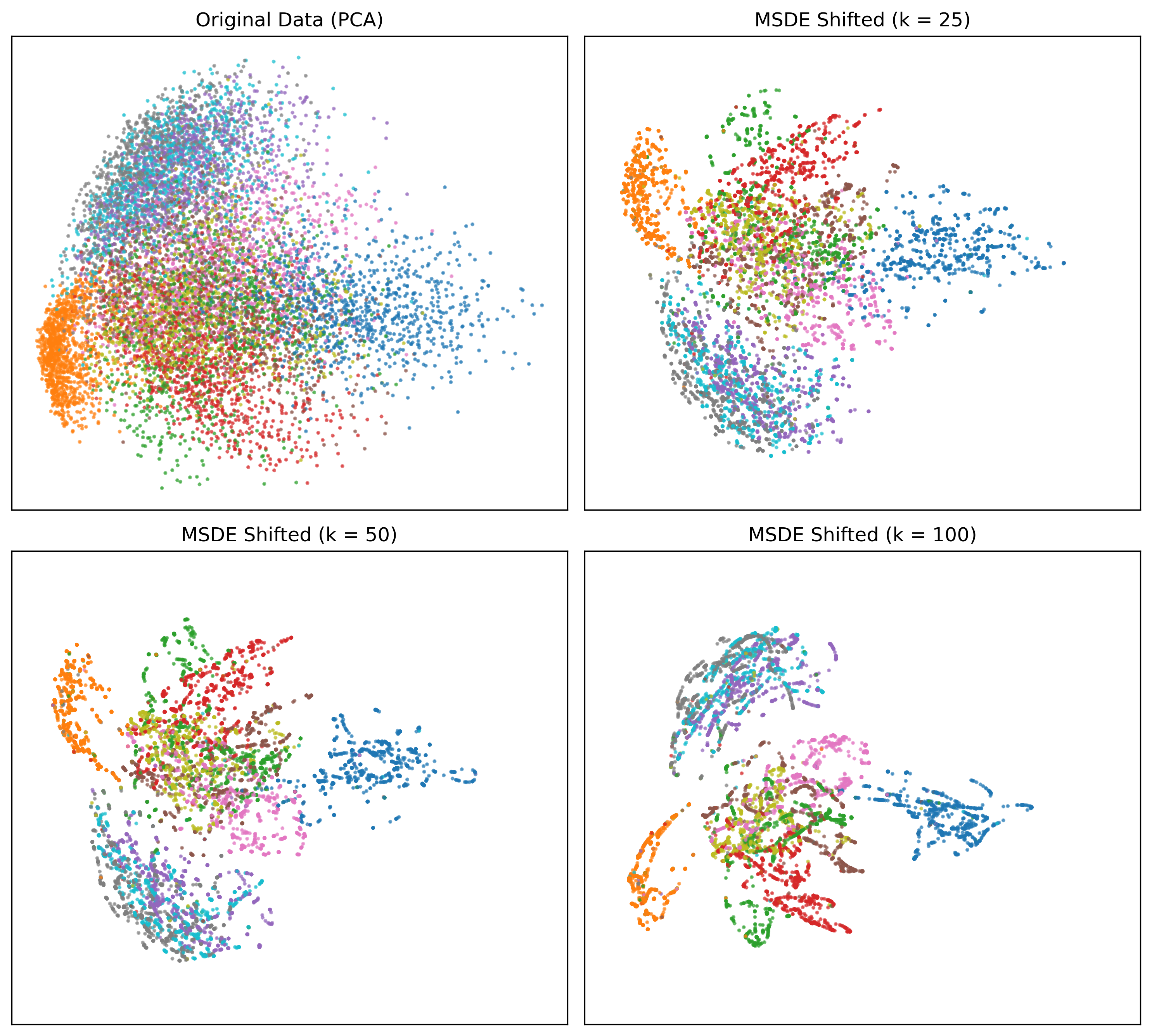}
    \caption{
    Geometric effect of Mean Shift Density Enhancement (MSDE) visualized in a two-dimensional PCA projection.
    \textbf{Top-left:} original data distribution of MNIST dataset.
    \textbf{Top-right and bottom:} data after MSDE updates (on the MNIST dataset) with increasing neighborhood sizes ($k=25,50,100$) and learning rate of 1.
    As $k$ increases, points are progressively attracted toward stable, high-density manifold regions, while samples originating in low-density or unstable regions undergo larger, more directed displacements.
    This illustrates the central principle of MSDE: anomalous points exhibit greater geometric instability under density-driven manifold evolution.
    }
    \label{fig:msde_pca_shift}
\end{figure}

\begin{figure}[htbp!]
    \centering
    \includegraphics[width=0.7\linewidth]{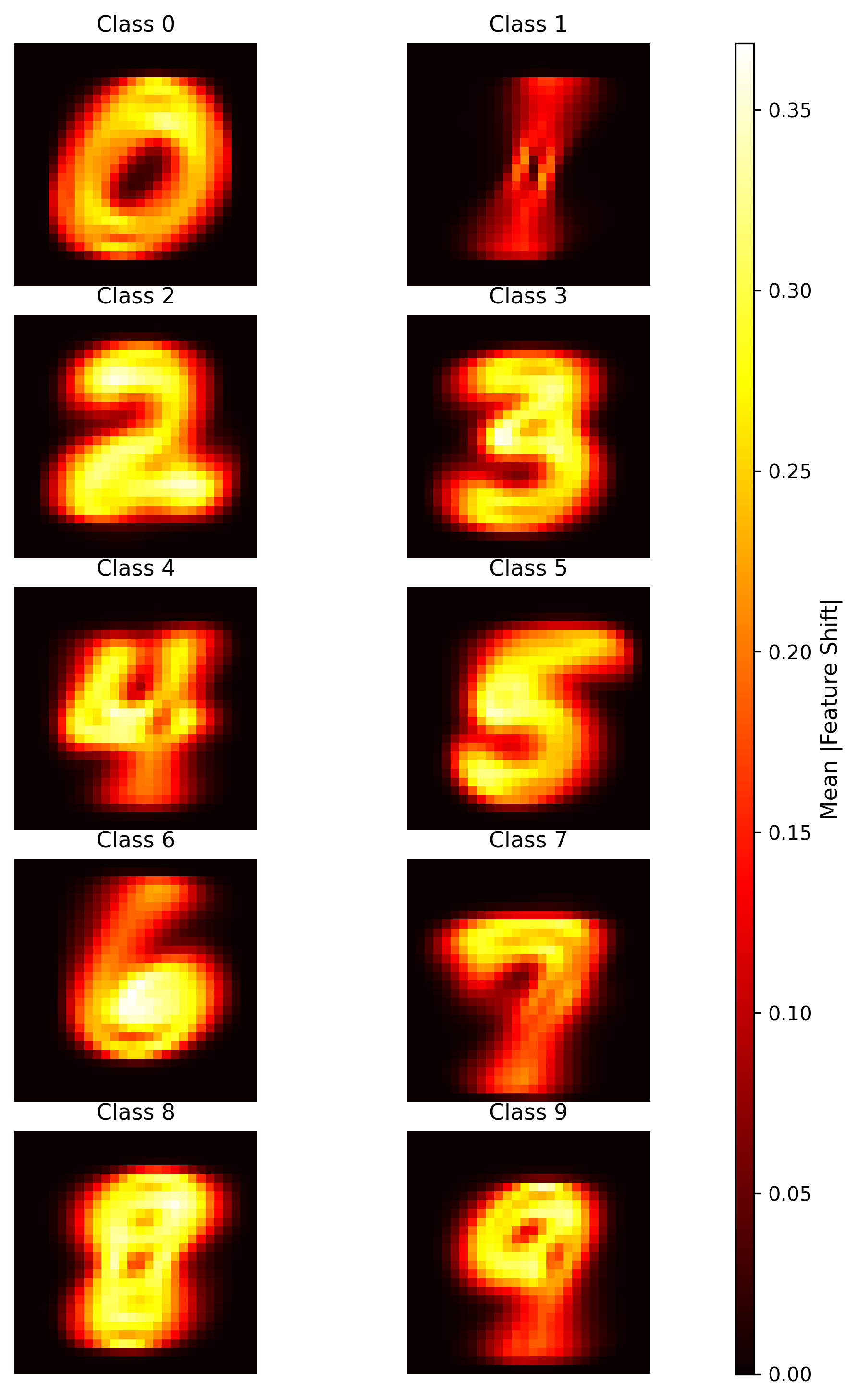}
    \caption{
    Class-wise visualization of mean absolute feature displacement induced by MSDE on the MNIST dataset.
    For each digit class, pixel intensities represent the average magnitude of per-feature shift accumulated during the mean-shift iterations.
    Displacement is concentrated along digit strokes and structurally informative regions, while background pixels remain largely stable.
    This indicates that MSDE-driven density enhancement selectively perturbs semantically meaningful features rather than introducing diffuse or global distortion.
    }
    \label{fig:mnist_feature_shift}
\end{figure}

\begin{longtable}{llccc}
\caption{Performance comparison of anomaly detection methods under varying noise levels.}\\

\toprule
Method & Noise & AUC-ROC $\uparrow$ & AUC-PR $\uparrow$ & Precision@n $\uparrow$ \\
\midrule
\endfirsthead

\toprule
Method & Noise & AUC-ROC $\uparrow$ & AUC-PR $\uparrow$ & Precision@n $\uparrow$ \\
\midrule
\endhead

\midrule
\multicolumn{5}{r}{\emph{Continued on next page}} \\
\endfoot

\bottomrule
\endlastfoot

MSDE & \multirow{13}{*}{0.00} &\textbf{0.921 ± 0.064} & \textbf{0.713 ± 0.236} & \textbf{0.695 ± 0.227} \\
CBLOF & &0.903 ± 0.061 & 0.689 ± 0.231 & 0.667 ± 0.223 \\
COF & &0.815 ± 0.179 & 0.568 ± 0.206 & 0.545 ± 0.199 \\
COPOD & &0.855 ± 0.159 & 0.624 ± 0.333 & 0.588 ± 0.337 \\
DAGMM & &0.822 ± 0.117 & 0.516 ± 0.219 & 0.499 ± 0.200 \\
ECOD & &0.852 ± 0.160 & 0.625 ± 0.315 & 0.585 ± 0.320 \\
HBOS & &0.853 ± 0.163 & 0.632 ± 0.337 & 0.600 ± 0.343 \\
IForest & &0.908 ± 0.088 & 0.666 ± 0.298 & 0.638 ± 0.293 \\
KNN & &0.910 ± 0.061 & 0.676 ± 0.190 & 0.655 ± 0.193 \\
LODA & &0.847 ± 0.130 & 0.612 ± 0.300 & 0.583 ± 0.290\\
LOF & &0.794 ± 0.210 & 0.574 ± 0.224 & 0.555 ± 0.217 \\
OCSVM & &0.856 ± 0.149 & 0.650 ± 0.332 & 0.621 ± 0.335 \\
PCA & &0.865 ± 0.154 & 0.641 ± 0.333 & 0.617 ± 0.339 \\
SOD & &0.877 ± 0.086 & 0.637 ± 0.195 & 0.617 ± 0.188 \\

\midrule
MSDE & \multirow{14}{*}{0.01} & \textbf{0.922 ± 0.064} & \textbf{0.713 ± 0.236} & \textbf{0.694 ± 0.227} \\
CBLOF & & 0.903 ± 0.061 & 0.687 ± 0.231 & 0.666 ± 0.223 \\
COF &  & 0.815 ± 0.179 & 0.567 ± 0.206 & 0.544 ± 0.198 \\
COPOD &  & 0.855 ± 0.160 & 0.624 ± 0.334 & 0.588 ± 0.337 \\
DAGMM &  & 0.823 ± 0.119 & 0.518 ± 0.224 & 0.501 ± 0.204 \\
ECOD &  & 0.852 ± 0.160 & 0.624 ± 0.315 & 0.584 ± 0.320 \\
HBOS &   & 0.853 ± 0.163 & 0.632 ± 0.337 & 0.599 ± 0.342 \\
IForest &  & 0.908 ± 0.088 & 0.666 ± 0.297 & 0.638 ± 0.292 \\
KNN &  & 0.910 ± 0.061 & 0.677 ± 0.189 & 0.655 ± 0.191 \\
LODA &  & 0.847 ± 0.130 & 0.613 ± 0.301 & 0.583 ± 0.291 \\
LOF &  & 0.795 ± 0.211 & 0.574 ± 0.225 & 0.555 ± 0.218 \\
OCSVM &  & 0.856 ± 0.149 & 0.650 ± 0.332 & 0.621 ± 0.335 \\
PCA &  & 0.865 ± 0.154 & 0.641 ± 0.333 & 0.616 ± 0.339 \\
SOD &  & 0.877 ± 0.087 & 0.636 ± 0.195 & 0.616 ± 0.189 \\

\midrule
MSDE & \multirow{14}{*}{0.05} & \textbf{0.921 ± 0.065} & \textbf{0.711 ± 0.240} & \textbf{0.690 ± 0.233} \\
CBLOF & & 0.900 ± 0.065 & 0.676 ± 0.242 & 0.653 ± 0.234 \\
COF &  & 0.819 ± 0.170 & 0.569 ± 0.203 & 0.545 ± 0.193 \\
COPOD &  & 0.854 ± 0.160 & 0.622 ± 0.334 & 0.586 ± 0.337 \\
DAGMM &  & 0.812 ± 0.121 & 0.498 ± 0.224 & 0.482 ± 0.202 \\
ECOD &  & 0.851 ± 0.161 & 0.623 ± 0.316 & 0.583 ± 0.319 \\
HBOS &  & 0.853 ± 0.163 & 0.630 ± 0.337 & 0.598 ± 0.343 \\
IForest &  & 0.906 ± 0.089 & 0.663 ± 0.299 & 0.635 ± 0.292 \\
KNN &  & 0.914 ± 0.055 & 0.690 ± 0.176 & 0.664 ± 0.181 \\
LODA &  & 0.846 ± 0.131 & 0.611 ± 0.301 & 0.582 ± 0.290 \\
LOF &  & 0.801 ± 0.204 & 0.574 ± 0.229 & 0.556 ± 0.221 \\
OCSVM &  & 0.853 ± 0.152 & 0.645 ± 0.334 & 0.616 ± 0.337 \\
PCA &  & 0.865 ± 0.155 & 0.641 ± 0.334 & 0.610 ± 0.340 \\
SOD &  & 0.877 ± 0.086 & 0.636 ± 0.195 & 0.614 ± 0.190 \\

\midrule
MSDE & \multirow{11}{*}{0.10} & 0.917 ± 0.072 & \textbf{0.705 ± 0.253} & \textbf{0.686 ± 0.242} \\
CBLOF &  & 0.893 ± 0.076 & 0.661 ± 0.265 & 0.642 ± 0.253 \\
COF &  & 0.820 ± 0.158 & 0.552 ± 0.203 & 0.527 ± 0.189 \\
COPOD &  & 0.853 ± 0.161 & 0.621 ± 0.335 & 0.588 ± 0.336 \\
DAGMM &  & 0.781 ± 0.127 & 0.455 ± 0.212 & 0.445 ± 0.193 \\
ECOD &  & 0.850 ± 0.161 & 0.623 ± 0.317 & 0.583 ± 0.320 \\
HBOS &  & 0.853 ± 0.163 & 0.631 ± 0.338 & 0.600 ± 0.342 \\
IForest &  & 0.904 ± 0.093 & 0.659 ± 0.303 & 0.630 ± 0.297 \\
KNN &  & \textbf{0.919 ± 0.049} & 0.702 ± 0.175 & 0.678 ± 0.176 \\
LODA &  & 0.841 ± 0.135 & 0.600 ± 0.303 & 0.570 ± 0.293 \\
LOF &  & 0.806 ± 0.193 & 0.577 ± 0.228 & 0.557 ± 0.221 \\
OCSVM &  & 0.842 ± 0.157 & 0.635 ± 0.336 & 0.607 ± 0.336 \\
PCA &  & 0.864 ± 0.156 & 0.640 ± 0.334 & 0.614 ± 0.339 \\
SOD &  & 0.871 ± 0.090 & 0.621 ± 0.206 & 0.599 ± 0.201 \\

\midrule
MSDE & \multirow{14}{*}{0.25} & 0.895 ± 0.092 & 0.681 ± 0.282 & 0.657 ± 0.269 \\
CBLOF &  & 0.875 ± 0.099 & 0.658 ± 0.276 & 0.634 ± 0.262 \\
COF &  & 0.804 ± 0.131 & 0.505 ± 0.207 & 0.481 ± 0.189 \\
COPOD &  & 0.847 ± 0.164 & 0.614 ± 0.331 & 0.578 ± 0.328 \\
DAGMM &  & 0.706 ± 0.135 & 0.350 ± 0.171 & 0.347 ± 0.166 \\
ECOD &  & 0.847 ± 0.163 & 0.618 ± 0.314 & 0.579 ± 0.317 \\
HBOS &  & 0.854 ± 0.163 & 0.636 ± 0.340 & 0.605 ± 0.345 \\
IForest &  & 0.893 ± 0.104 & 0.653 ± 0.318 & 0.623 ± 0.316 \\
KNN &  & \textbf{0.912 ± 0.069} & \textbf{0.704 ± 0.227} & \textbf{0.678 ± 0.218} \\
LODA &  & 0.834 ± 0.148 & 0.586 ± 0.311 & 0.555 ± 0.299 \\
LOF &  & 0.808 ± 0.163 & 0.561 ± 0.227 & 0.541 ± 0.217 \\
OCSVM &  & 0.820 ± 0.166 & 0.608 ± 0.336 & 0.575 ± 0.331 \\
PCA &  & 0.862 ± 0.156 & 0.642 ± 0.335 & 0.615 ± 0.339 \\
SOD &  & 0.852 ± 0.097 & 0.601 ± 0.209 & 0.579 ± 0.203 \\

\midrule
MSDE & \multirow{14}{*}{0.50} & 0.852 ± 0.126 & 0.619 ± 0.315 & 0.593 ± 0.300 \\
CBLOF & & 0.854 ± 0.104 & 0.636 ± 0.267 & 0.611 ± 0.256 \\
COF &  & 0.770 ± 0.127 & 0.462 ± 0.220 & 0.439 ± 0.195 \\
COPOD &  & 0.837 ± 0.165 & 0.597 ± 0.325 & 0.560 ± 0.321 \\
DAGMM &  & 0.639 ± 0.123 & 0.247 ± 0.115 & 0.252 ± 0.122 \\
ECOD &  & 0.838 ± 0.161 & 0.596 ± 0.304 & 0.557 ± 0.303 \\
HBOS &  & 0.855 ± 0.160 & 0.630 ± 0.337 & 0.598 ± 0.343 \\
IForest &  & 0.871 ± 0.127 & 0.617 ± 0.321 & 0.588 ± 0.317 \\
KNN &  & \textbf{0.872 ± 0.111} & \textbf{0.652 ± 0.292} & \textbf{0.625 ± 0.279} \\
LODA &  & 0.800 ± 0.157 & 0.514 ± 0.304 & 0.486 ± 0.284 \\
LOF &  & 0.783 ± 0.144 & 0.524 ± 0.246 & 0.502 ± 0.233 \\
OCSVM &  & 0.796 ± 0.169 & 0.557 ± 0.323 & 0.524 ± 0.313 \\
PCA &  & 0.855 ± 0.159 & 0.632 ± 0.334 & 0.606 ± 0.338 \\
SOD &  & 0.852 ± 0.083 & 0.601 ± 0.193 & 0.579 ± 0.182 
\label{tab:noise_multimodel}
\end{longtable}

\begin{longtable}{llccc}
\caption{Performance comparison of anomaly detection methods under varying synthetic modes.}\\

\toprule
Method & Synthetic Modes & AUC-ROC $\uparrow$ & AUC-PR $\uparrow$ & Precision@n $\uparrow$ \\
\midrule
\endfirsthead

\toprule
Method & Synthetic Modes & AUC-ROC $\uparrow$ & AUC-PR $\uparrow$ & Precision@n $\uparrow$ \\
\midrule
\endhead

\midrule
\multicolumn{5}{r}{\emph{Continued on next page}} \\
\endfoot

\bottomrule
\endlastfoot
MSDE & \multirow{14}{*}{Dependency} & 0.834 ± 0.140 & 0.390 ± 0.280 & 0.389 ± 0.296 \\
MSDE (opt) & & 0.880 ± 0.124 & 0.521 ± 0.308 & 0.515 ± 0.300 \\
CBLOF & & 0.837 ± 0.133 & 0.376 ± 0.286 & 0.368 ± 0.310 \\
COF & & 0.886 ± 0.123 & \textbf{0.569 ± 0.312} & 0.553 ± 0.288 \\
COPOD & & 0.608 ± 0.084 & 0.149 ± 0.152 & 0.107 ± 0.173 \\
DAGMM & & 0.652 ± 0.129 & 0.211 ± 0.181 & 0.225 ± 0.190 \\
ECOD & & 0.597 ± 0.089 & 0.147 ± 0.151 & 0.098 ± 0.170 \\
HBOS & & 0.600 ± 0.086 & 0.147 ± 0.151 & 0.109 ± 0.175 \\
IForest & & 0.782 ± 0.128 & 0.251 ± 0.218 & 0.235 ± 0.266 \\
KNN & & \textbf{0.895 ± 0.118} & 0.567 ± 0.318 & \textbf{0.556 ± 0.313} \\
LODA & & 0.649 ± 0.116 & 0.172 ± 0.165 & 0.154 ± 0.191 \\
LOF & & 0.879 ± 0.131 & \textbf{0.569 ± 0.330} & \textbf{0.556 ± 0.307} \\
OCSVM & & 0.624 ± 0.113 & 0.163 ± 0.166 & 0.128 ± 0.185 \\
PCA & & 0.626 ± 0.107 & 0.169 ± 0.173 & 0.135 ± 0.188 \\
SOD & & 0.893 ± 0.115 & 0.542 ± 0.303 & 0.532 ± 0.291 \\

\midrule
MSDE & \multirow{14}{*}{Cluster} & 0.952 ± 0.094 & 0.886 ± 0.190 & 0.861 ± 0.202 \\
MSDE (opt)& & 0.944 ± 0.101 & 0.868 ± 0.201 &  0.841 ± 0.211   \\
CBLOF & & 0.889 ± 0.163 & 0.828 ± 0.240 & 0.802 ± 0.254 \\
COF & & 0.521 ± 0.245 & 0.271 ± 0.283 & 0.254 ± 0.278 \\
COPOD & & 0.969 ± 0.066 & 0.927 ± 0.142 & 0.900 ± 0.170 \\
DAGMM & & 0.931 ± 0.112 & 0.732 ± 0.229 & 0.704 ± 0.223 \\
ECOD & & 0.947 ± 0.098 & 0.855 ± 0.190 & 0.819 ± 0.207 \\
HBOS & & 0.970 ± 0.082 & 0.928 ± 0.167 & 0.909 ± 0.183 \\
IForest & & 0.969 ± 0.079 & 0.918 ± 0.153 & 0.886 ± 0.180 \\
KNN & & 0.835 ± 0.141 & 0.503 ± 0.206 & 0.475 ± 0.184 \\
LODA & & 0.917 ± 0.178 & 0.827 ± 0.259 & 0.793 ± 0.271 \\
LOF & & 0.447 ± 0.214 & 0.229 ± 0.250 & 0.218 ± 0.250 \\
OCSVM & & 0.956 ± 0.102 & \textbf{0.941 ± 0.139} & 0.919 ± 0.161 \\
PCA & & \textbf{0.979 ± 0.066} & 0.936 ± 0.139 & \textbf{0.929 ± 0.139} \\
SOD & & 0.758 ± 0.196 & 0.474 ± 0.297 & 0.457 ± 0.276 \\

\midrule
MSDE & \multirow{14}{*}{Global} & 0.998 ± 0.005 & 0.958 ± 0.132 & 0.935 ± 0.143 \\
MSDE (opt) & & 0.998 ± 0.004 & 0.958 ± 0.133 & 0.937 ± 0.143 \\
CBLOF & & 0.997 ± 0.007 & 0.952 ± 0.136 & 0.924 ± 0.151 \\
COF & & 0.945 ± 0.116 & 0.824 ± 0.256 & 0.788 ± 0.267 \\
COPOD & & 0.991 ± 0.019 & 0.904 ± 0.175 & 0.865 ± 0.193 \\
DAGMM & & 0.913 ± 0.077 & 0.688 ± 0.191 & 0.649 ± 0.168 \\
ECOD & & 0.990 ± 0.018 & 0.910 ± 0.179 & 0.871 ± 0.195 \\
HBOS & & 0.993 ± 0.015 & 0.922 ± 0.166 & 0.887 ± 0.186 \\
IForest & & 0.997 ± 0.005 & 0.943 ± 0.158 & 0.913 ± 0.168 \\
KNN & & \textbf{0.999 ± 0.003} & \textbf{0.975 ± 0.109} & \textbf{0.960 ± 0.134} \\
LODA & & 0.979 ± 0.074 & 0.904 ± 0.169 & 0.862 ± 0.176 \\
LOF & & 0.916 ± 0.169 & 0.788 ± 0.295 & 0.758 ± 0.314 \\
OCSVM & & 0.987 ± 0.061 & 0.925 ± 0.155 & 0.894 ± 0.159 \\
PCA & & 0.993 ± 0.017 & 0.930 ± 0.144 & 0.899 ± 0.158 \\
SOD & & 0.989 ± 0.030 & 0.953 ± 0.098 & 0.922 ± 0.122 \\

\midrule
MSDE & \multirow{14}{*}{Local} & 0.904 ± 0.096 & 0.620 ± 0.318 & 0.593 ± 0.300 \\
MSDE (opt) & & 0.912 ± 0.092 & 0.635 ± 0.312 & 0.608 ± 0.298  \\
CBLOF & & 0.891 ± 0.112 & 0.599 ± 0.324 & 0.575 ± 0.307 \\
COF & & 0.907 ± 0.080 & 0.609 ± 0.291 & 0.585 ± 0.261 \\
COPOD & & 0.853 ± 0.118 & 0.516 ± 0.309 & 0.482 ± 0.283 \\
DAGMM & & 0.791 ± 0.102 & 0.433 ± 0.242 & 0.419 ± 0.211 \\
ECOD & & 0.875 ± 0.114 & 0.587 ± 0.304 & 0.552 ± 0.279 \\
HBOS & & 0.850 ± 0.128 & 0.529 ± 0.313 & 0.497 ± 0.287 \\
IForest & & 0.885 ± 0.095 & 0.556 ± 0.312 & 0.520 ± 0.290 \\
KNN & & 0.912 ± 0.095 & 0.661 ± 0.309 & 0.627 ± 0.289 \\
LODA & & 0.8433 ± 0.136 & 0.546 ± 0.300 & 0.523 ± 0.271 \\
LOF & & \textbf{0.936 ± 0.075} & \textbf{0.710 ± 0.288} & \textbf{0.689 ± 0.263} \\
OCSVM & & 0.858 ± 0.131 & 0.571 ± 0.317 & 0.544 ± 0.293 \\
PCA & & 0.864 ± 0.122 & 0.529 ± 0.309 & 0.503 ± 0.291 \\
SOD & & 0.870 ± 0.101 & 0.580 ± 0.290 & 0.558 ± 0.263 
\label{tab:syntheticmode_multimodel}
\end{longtable}

\paragraph{Effect of hyperparameter optimization.}
In addition to the default MSDE configuration, we report results for \textbf{MSDE (opt)}, where hyperparameters are tuned separately for each synthetic mode using Optuna with 10 optimization trials. As shown in Table~\ref{tab:syntheticmode_multimodel}, MSDE (opt) consistently improves performance over the default configuration under \emph{Dependency}, \emph{Global}, and \emph{Local} synthetic modes across all evaluation metrics. 

For the \emph{Cluster} mode, MSDE (opt) exhibits slightly lower performance than the default MSDE. We attribute this behavior to the limited optimization budget: with only 10 trials, the hyperparameter search may not sufficiently explore configurations that favor highly separable, globally compact cluster structures. In such settings, the default MSDE parameters designed to emphasize conservative neighborhood aggregation already align well with the underlying data geometry. Increasing the number of optimization trials or adopting mode-specific search spaces is expected to close this gap.

Overall, these results demonstrate that MSDE is robust to hyperparameter choices, while targeted optimization can further enhance performance in most anomaly regimes.

\begin{table}[htbp!]
\centering
\caption{Optimized MSDE hyperparameters obtained via Optuna for each synthetic anomaly mode.}
\label{tab:msde_opt_params}
\begin{tabular}{lcccccc}
\hline
\textbf{Synthetic Mode} 
& $k$ 
& $T_{\text{nbd}}$ 
& $\eta$ 
& $T$ 
& $\tau$ 
& $\theta$ \\
\hline
Dependency 
& 13 
& 35 
& 0.0945 
& 19 
& $2.25 \times 10^{-4}$ 
& 0.294 \\
\hline
Cluster 
& 82 
& 67 
& 0.1171 
& 7 
& $2.69 \times 10^{-4}$ 
& 0.138 \\
\hline
Global 
& 61 
& 70 
& 0.1210 
& 20 
& $1.19 \times 10^{-4}$ 
& 0.115 \\
\hline
Local 
& 26 
& 32 
& 0.1303 
& 13 
& $2.98 \times 10^{-5}$ 
& 0.045 \\
\hline
\end{tabular}
\end{table}

\paragraph{Dataset-wise Ranking and Robustness Analysis.}
The dataset-wise evaluation averaged across all four synthetic anomaly generation modes further demonstrates the strong and consistent performance of MSDE across diverse anomaly detection scenarios. As shown in Table~\ref{tab:overall_results}, MSDE appears among the top three performing methods in 37 out of the 46 benchmark datasets, highlighting its robustness and generalization capability across datasets with varying characteristics and anomaly structures. In particular, MSDE achieves the first rank on a substantial number of datasets, while also consistently maintaining competitive second- and third-place positions when not ranked first. This indicates that the proposed approach does not rely on exceptional performance on only a small subset of datasets, but instead delivers stable and reliable performance across a broad spectrum of anomaly detection tasks. Furthermore, the reported mean $\pm$ standard deviation values provide additional evidence of the stability and consistency of MSDE across multiple experimental runs and synthetic anomaly settings.

\begin{table*}[t]
\centering
\caption{per-dataset average performance comparison (AUC-ROC) of anomaly detection methods across all four synthetic anomaly generation modes under zero noise setting. The full results are available in the supplementary file ``\texttt{overall\_aucroc\_rankings.csv}"}
\label{tab:overall_results}
\resizebox{\textwidth}{!}{
\begin{tabular}{lccc}
\toprule
\textbf{Dataset} & \textbf{1st Rank} & \textbf{2nd Rank} & \textbf{3rd Rank} \\
\midrule

annthyroid       & KNN (0.897 ± 0.069) & \textbf{MSDE (0.896 ± 0.098)} & IForest (0.880 ± 0.148) \\
backdoor         & PCA (1.000 ± 0.000) & \textbf{MSDE (1.000 ± 0.001)} & LODA (0.997 ± 0.005)\\
breastw          & LODA (0.840 ± 0.195) & PCA (0.839 ± 0.188) & \textbf{MSDE (0.836 ± 0.139)} \\
campaign         & \textbf{MSDE (0.977 ± 0.039)} & CBLOF (0.966 ± 0.054) & PCA (0.948 ± 0.099) \\
cardio           & CBLOF (0.966 ± 0.034) & \textbf{MSDE (0.966 ± 0.035)} &KNN (0.946 ± 0.054) \\
Cardiotocography & \textbf{MSDE (0.956 ± 0.049)} & IForest (0.931 ± 0.098) & KNN (0.907 ± 0.126) \\
celeba           & \textbf{MSDE (0.992 ± 0.015)} & KNN (0.975 ± 0.032) & CBLOF (0.971 ± 0.057) \\
census           & CBLOF (0.976 ± 0.046) & \textbf{MSDE (0.957 ± 0.055)} & IForest (0.945 ± 0.042) \\
cover            & KNN (0.977 $\pm$ 0.033) & \textbf{MSDE (0.967 $\pm$ 0.039)} & SOD (0.963 $\pm$ 0.055) \\
donors           & \textbf{MSDE (0.919 $\pm$ 0.094)} & CBLOF (0.896 $\pm$ 0.120) & IForest (0.885 $\pm$ 0.133) \\
fault            & \textbf{MSDE (0.976 $\pm$ 0.033)} & IForest (0.962 $\pm$ 0.046) & PCA (0.949 $\pm$ 0.058) \\
fraud            & KNN (0.984 $\pm$ 0.028) & \textbf{MSDE (0.981 $\pm$ 0.031)} & CBLOF (0.980 $\pm$ 0.027) \\
glass            & COF (0.911 $\pm$ 0.068) & SOD (0.904 $\pm$ 0.104) & CBLOF (0.886 $\pm$ 0.132) \\
Hepatitis        & IForest (0.919 $\pm$ 0.115) & \textbf{MSDE (0.919 $\pm$ 0.114)} & KNN (0.911 $\pm$ 0.063) \\
http             & \textbf{MSDE (0.837 $\pm$ 0.209)} & ECOD (0.836 $\pm$ 0.213) & IForest (0.835 $\pm$ 0.216) \\
InternetAds      & COPOD (0.915 $\pm$ 0.147) & DAGMM (0.907 $\pm$ 0.083) & PCA (0.898 $\pm$ 0.177) \\
Ionosphere       & \textbf{MSDE (0.910 $\pm$ 0.101)} & SOD (0.910 $\pm$ 0.126) & CBLOF (0.907 $\pm$ 0.103) \\
landsat          & IForest (0.943 $\pm$ 0.071) & \textbf{MSDE (0.938 $\pm$ 0.081)} & KNN (0.897 $\pm$ 0.159) \\
letter           & CBLOF (0.988 $\pm$ 0.014) & \textbf{MSDE (0.987 $\pm$ 0.016)} & SOD (0.984 $\pm$ 0.029) \\
Lymphography     & KNN (0.971 ± 0.031) & CBLOF (0.965 ± 0.044) & COF (0.963 ± 0.039)\\
magic.gamma      & \textbf{MSDE (0.894 ± 0.098)} & IForest (0.890 ± 0.126) & KNN (0.888 ± 0.158)\\
mammography      & \textbf{MSDE (0.881 ± 0.147)} & IForest (0.863 ± 0.161) & CBLOF (0.843 ± 0.183) \\
mnist            & OCSVM (1.000 ± 0.000) & ECOD (1.000 ± 0.000) & COPOD (1.000 ± 0.000)\\
musk             & \textbf{MSDE (0.998 ± 0.002)} & SOD (0.994 ± 0.005) & KNN (0.994 ± 0.009)\\
optdigits        & \textbf{MSDE (0.999 ± 0.002)} & ECOD (0.999 ± 0.002) & OCSVM (0.997 ± 0.005)\\
PageBlocks       & \textbf{MSDE (0.850 ± 0.149)} & SOD (0.811 ± 0.133) & IForest (0.811 ± 0.214) \\
pendigits        & \textbf{MSDE (0.983 ± 0.027)} & SOD (0.972 ± 0.044) & CBLOF (0.971 ± 0.045) \\
Pima             & \textbf{MSDE (0.880 ± 0.173)} & IForest (0.862 ± 0.172) & KNN (0.854 ± 0.134)\\
satellite        & IForest (0.952 ± 0.056) & \textbf{MSDE (0.939 ± 0.088)} & CBLOF (0.894 ± 0.142)\\
satimage-2       & \textbf{MSDE (0.989 ± 0.020)} & SOD (0.986 ± 0.015) & CBLOF (0.978 ± 0.037) \\
shuttle          & KNN (0.924 ± 0.051) & IForest (0.919 ± 0.096) & \textbf{MSDE (0.909 ± 0.06)} \\
skin             & IForest (0.785 ± 0.210) & \textbf{MSDE (0.775 ± 0.141)} & KNN (0.758 ± 0.152) \\
smtp             & SOD (0.926 ± 0.086) & KNN (0.854 ± 0.190) & IForest (0.847 ± 0.201) \\
SpamBase         & \textbf{MSDE (0.848 ± 0.170)} & CBLOF (0.842 ± 0.179) & KNN (0.838 ± 0.194) \\
speech           & \textbf{MSDE (0.998 ± 0.003)} & CBLOF (0.993 ± 0.014) & KNN (0.993 ± 0.014)\\
Stamps           & CBLOF (0.921 ± 0.087) & KNN (0.914 ± 0.062) & \textbf{MSDE (0.911 ± 0.108)} \\
thyroid          & KNN (0.893 ± 0.093) & \textbf{MSDE (0.878 ± 0.144)} & SOD (0.876 ± 0.104) \\
vertebral        & KNN (0.917 ± 0.056) & \textbf{MSDE (0.897 ± 0.123)} & IForest (0.893 ± 0.125) \\
vowels           & LOF (0.988 ± 0.02) & COF (0.981 ± 0.028) & KNN (0.980 ± 0.03) \\
Waveform         & \textbf{MSDE (0.993 ± 0.01)} & KNN (0.991 ± 0.011) & CBLOF (0.985 ± 0.026) \\
WBC              & SOD (0.895 ± 0.093) & KNN (0.890 ± 0.082) & \textbf{MSDE (0.882 ± 0.146)} \\
WDBC             & KNN (0.995 ± 0.006) & CBLOF (0.991 ± 0.010) & SOD (0.988 ± 0.018) \\
Wilt             & KNN (0.854 ± 0.136) & \textbf{MSDE (0.854 ± 0.184)} & CBLOF (0.840 ± 0.193)\\
wine             & KNN (0.965 ± 0.030) & CBLOF (0.962 ± 0.044) & SOD (0.948 ± 0.037)\\
WPBC             & \textbf{MSDE (0.945 ± 0.066)} & IForest (0.927 ± 0.08) & CBLOF (0.922 ± 0.068)\\
yeast            & IForest (0.904 ± 0.084) & KNN (0.878 ± 0.125) & ECOD (0.875 ± 0.134)\\

\bottomrule
\end{tabular}
}
\end{table*}

\begin{figure*}[htbp!]
    \centering
    \includegraphics[width=\linewidth]{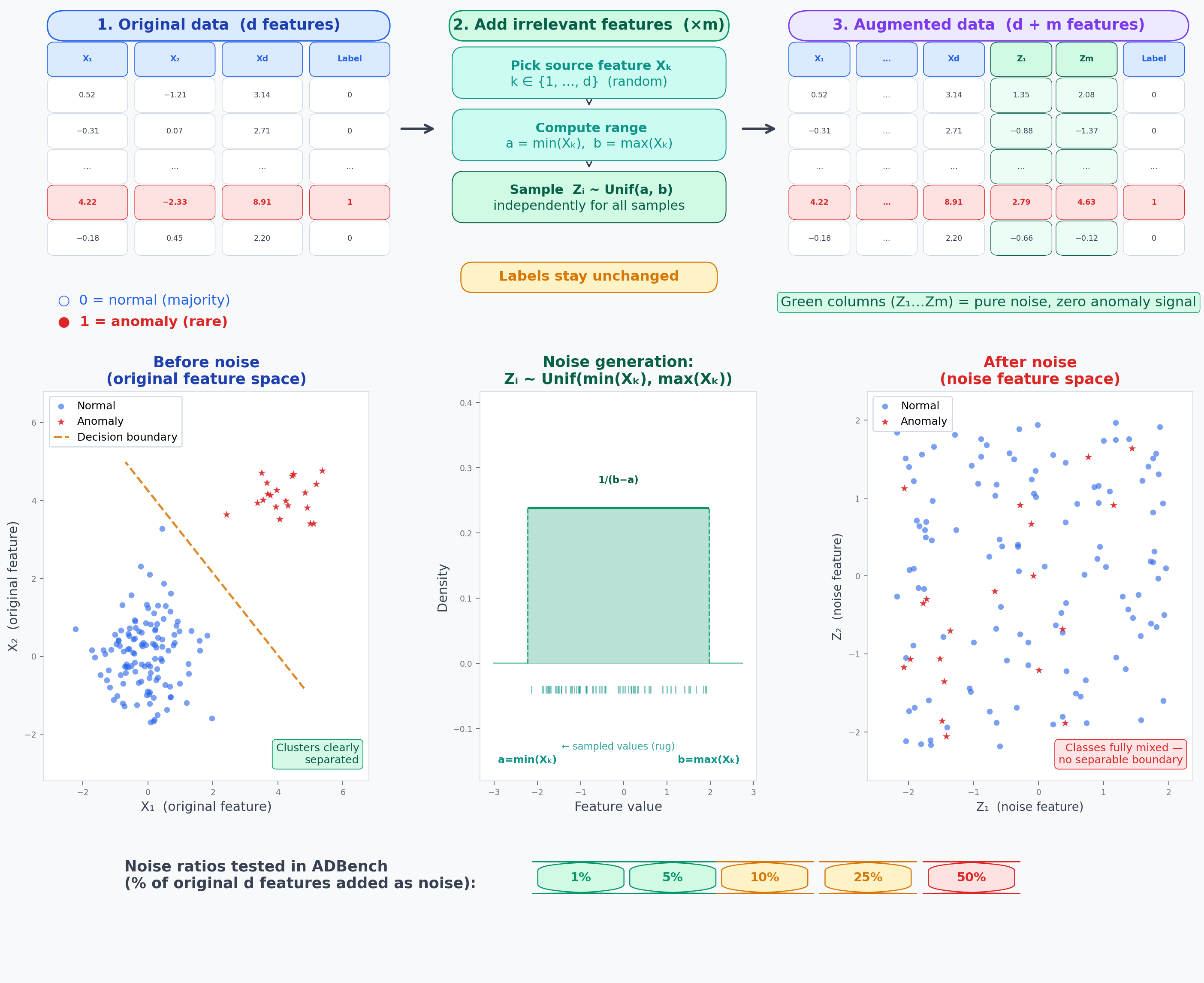}
    \caption{Illustration of the irrelevant feature noise simulation strategy used in ADBench. Starting from the original feature space, additional synthetic dimensions are generated by randomly selecting an existing feature $X_k$ and sampling new values from a uniform distribution $Z_i \sim \mathrm{Uniform}(\min(X_k), \max(X_k))$. The injected features preserve the numerical range of the data but contain no anomaly-discriminative signal, while class labels remain unchanged. As the number of irrelevant features increases, the feature space becomes increasingly noisy, masking the underlying anomaly structure relied upon by unsupervised anomaly detectors and leading to performance degradation (reported up to $\sim10\%$ AUC-ROC reduction at $50\%$ noise in ADBench).}
    \label{fig:irrelevant_feature_noise}
\end{figure*}

\paragraph{Robustness to Irrelevant Feature Noise in ADBench}
To evaluate robustness against high-dimensional irrelevant information, we follow the \textit{irrelevant feature noise} protocol introduced in ADBench. In this setting, synthetic noise dimensions are appended to the original feature space while preserving the original class labels. As illustrated in Figure~\ref{fig:irrelevant_feature_noise}, the added features contain no anomaly-discriminative information and therefore act as purely irrelevant dimensions.

Such corruption increases the effective dimensionality of the data and progressively obscures the intrinsic anomaly structure relied upon by unsupervised detectors. As the proportion of irrelevant features grows, anomalies become harder to separate from normal samples in the augmented feature space. ADBench reports that unsupervised methods may experience performance degradation of up to $\sim10\%$ AUC-ROC at the highest noise level ($50\%$ additional irrelevant features). Following the ADBench protocol, we evaluate noise ratios ranging from $1\%$ to $50\%$ of the original feature dimensionality.

\section{Scalability Experiments}\label{app:scalability}

\subsection{Dataset Description}
We performed a small-scale scalability analysis of the proposed \textbf{(MSDE)} method on the Credit Card Fraud Detection dataset~\citep{creditFraud} released by ULB. The dataset contains \textbf{284,807 real-world credit card transactions}, among which only \textbf{492 are fraudulent} (approximately \textbf{0.17\%}), making it extremely imbalanced. All features are numerical and anonymized through PCA transformation, except for the transaction amount and timestamp. Due to its large sample size, severe class imbalance, and frequent adoption in anomaly detection literature, this dataset serves as a challenging and realistic benchmark for scalability analysis.

\subsection{Experimental Setup}
All experiments were conducted using the \emph{ADBench} framework to ensure a standardized evaluation protocol across methods. MSDE was compared against a diverse set of unsupervised anomaly detection models, including density-based, distance-based, ensemble-based, and probabilistic approaches.

We report the \textbf{average total runtime}, which includes both model fitting and inference time. To enable fair comparison across models with substantially different computational characteristics, runtimes are reported on a \textbf{logarithmic scale}.

\subsection{Runtime Scalability Analysis}
Figure~\ref{fig:runtime_creditcard} presents the average total runtime of all evaluated models on the Credit Card Fraud dataset.


MSDE demonstrates \textbf{moderate computational cost}, outperforming several computationally intensive models such as KNN, DAGMM, SOD, and OCSVM, while remaining more expensive than lightweight linear or histogram-based methods (e.g., PCA and HBOS). The \textbf{COF} model failed to complete execution due to excessive memory consumption. As a result, runtime and performance metrics for COF are not reported. Distance-based neighborhood methods exhibit a sharp increase in runtime, highlighting their limited scalability on large datasets.



\subsection{Detection Performance Comparison}

\begin{table}[htbp!]
\centering
\caption{Performance and Runtime comparison on the Credit Card Fraud dataset.}
\label{tab:creditcard_performance}
\begin{tabular}{lcccc}
\toprule
\textbf{Model} & \textbf{AUC-ROC} & \textbf{AUC-PR} & \textbf{Precision@N} & \textbf{Avg. Total Runtime (sec)} \\
\midrule
MSDE     & \textbf{0.965 ± 0.004} & 0.289 ± 0.049 & 0.354 ± 0.049 & 281.013\\
PCA      & 0.959 ± 0.004 & 0.173 ± 0.018 & 0.250 ± 0.034 & \textbf{0.323}\\
ECOD     & 0.958 ± 0.002 & 0.250 ± 0.027 & 0.331 ± 0.029 & 3.984\\
HBOS     & 0.958 ± 0.009 & 0.253 ± 0.009 & 0.340 ± 0.010 & 1.256\\
COPOD    & 0.956 ± 0.002 & 0.281 ± 0.017 & 0.354 ± 0.022 & 3.973\\
IForest  & 0.955 ± 0.003 & 0.160 ± 0.077 & 0.232 ± 0.091 & 2.521\\
CBLOF    & 0.952 ± 0.003 & \textbf{0.435 ± 0.065} & 0.475 ± 0.055 & 1.056 \\
KNN      & 0.949 ± 0.006 & 0.150 ± 0.021 & 0.207 ± 0.024 & 1021.669\\
OCSVM    & 0.923 ± 0.004 & 0.390 ± 0.068 & 0.471 ± 0.040 & 5483.390\\
SOD      & 0.919 ± 0.004 & 0.048 ± 0.020 & 0.119 ± 0.038 & 4532.995\\
DAGMM    & 0.898 ± 0.076 & 0.431 ± 0.188 & \textbf{0.502 ± 0.176} & 2149.945\\
LODA     & 0.860 ± 0.069 & 0.262 ± 0.085 & 0.367 ± 0.032 & 3.183\\
LOF      & 0.511 ± 0.043 & 0.002 ± 0.000 & 0.007 ± 0.000 & 61.675\\
COF      & NA     & NA    & NA & NA\\
\bottomrule
\end{tabular}
\end{table}

Table~\ref{tab:creditcard_performance} summarizes the average total runtime (in seconds), and mean detection performance of all models in terms of AUC-ROC, AUC-PR, and Precision@N, which collectively capture ranking quality and detection effectiveness under severe class imbalance.

\subsection{Hardware and Software Environment.}
All scalability experiments for MSDE were conducted on a desktop system equipped with a 13th Gen Intel(R) Core(TM) i7-13700 processor (base frequency 2.10\,GHz), 32\,GB RAM (31.7\,GB usable), running Windows 11 (64-bit) on an x64-based architecture. 

The implementation was developed and executed using Python version 3.13.5. All experiments were performed on CPU without GPU acceleration. Unless otherwise stated, default multi-threading behavior of underlying numerical libraries (e.g., NumPy) was used. No distributed computing framework was employed.

\subsection{Discussion}

As shown in Table~\ref{tab:creditcard_performance}, lightweight linear and histogram-based approaches such as PCA and HBOS achieve extremely low runtimes, confirming their suitability for large-scale deployment. MSDE incurs a higher computational cost due to its graph-based construction and iterative optimization, yet remains substantially more efficient than distance-heavy methods such as KNN, DAGMM, SOD, and OCSVM. These neighborhood-based and kernel-based models exhibit orders-of-magnitude increases in runtime, underscoring their limited scalability on large, high-dimensional datasets.

MSDE achieves the \textbf{highest AUC-ROC} among all evaluated methods, indicating superior overall ranking capability, while maintaining competitive AUC-PR and Precision@N compared to strong precision-oriented baselines such as DAGMM and CBLOF.  

Importantly, MSDE offers a \textbf{favorable trade-off between detection performance and computational scalability}, making it well-suited for large-scale, real-world fraud detection scenarios. The failure of COF further emphasizes the necessity of memory-efficient algorithmic design when handling datasets with hundreds of thousands of samples.

\section{Hyperparameter Importance Analysis}
\label{app:hyperparam_importance}

\paragraph{Objective}
To better understand the sensitivity of the proposed MSDE framework to its design parameters, we analyze the relative importance of individual hyperparameters using Optuna's built-in hyperparameter importance estimation. This analysis aims to identify which components most strongly influence anomaly detection performance and to provide insights into the dominant factors governing the behavior of MSDE.

\paragraph{Experimental Setup.}
Hyperparameter optimization is performed using Optuna under the full ADBench evaluation protocol on the Credit Card Fraud Detection dataset. The objective function maximizes the mean AUC-ROC across runs. After completing the study, hyperparameter importances are computed using Optuna’s \texttt{get\_param\_importances} method, which estimates the contribution of each parameter to the objective using fANOVA~\citep{fanova}. fANOVA estimates the percentage of variance in the classification performance on the validation set explained by each hyperparameter, given a regression tree and a hyperparameter space. The larger the percentage, the greater impact on the classification performance on the validation set.

\paragraph{Results}
Table~\ref{tab:hyperparam_importance} reports the normalized importance scores for each hyperparameter, and Figure~\ref{fig:hyperparam_importance} visualizes the same information.

\begin{table}[htbp!]
\centering
\caption{Hyperparameter importance scores obtained from Optuna for the MSDE model on the Credit Card dataset. Higher values indicate a stronger influence on AUC-ROC performance.}
\label{tab:hyperparam_importance}
\begin{tabular}{lc}
\toprule
\textbf{Hyperparameter} & \textbf{Importance} \\
\midrule
Neighborhood count threshold ($T_{\text{nbd}}$) & 0.645 \\
Anomaly threshold ($\theta$) & 0.170 \\
Number of neighbors ($k$) & 0.113 \\
Learning rate ($\eta$) & 0.046 \\
Maximum shift iterations ($T$) & 0.022 \\
Shift threshold ($\tau$) & 0.004 \\
\bottomrule
\end{tabular}
\end{table}

\paragraph{Discussion}
The results indicate that the neighborhood sample count threshold ($T_{\text{nbd}}$), which governs the empirical density estimation, is by far the most influential hyperparameter. This highlights the central role of local density estimation in driving the mean-shift dynamics and anomaly discrimination capability of MSDE.

The anomaly threshold ($\theta$) and neighborhood size ($k$) form a secondary group of influential parameters, reflecting their role in translating accumulated shift behavior into anomaly scores and defining the locality of the underlying graph structure.

In contrast, optimization-related parameters such as the learning rate, maximum number of shift iterations, and convergence threshold exhibit relatively low importance. This suggests that the MSDE framework is robust to the precise tuning of its iterative dynamics, provided that density estimation and neighborhood structure are well-specified.

\paragraph{Effect of Varying Neighborhood Size \(k\) for Cluster Anomalies}

\begin{figure}[htbp!]
    \centering

    \begin{subfigure}{\textwidth}
        \centering
        \includegraphics[width=1\linewidth, height=4.8cm, keepaspectratio]{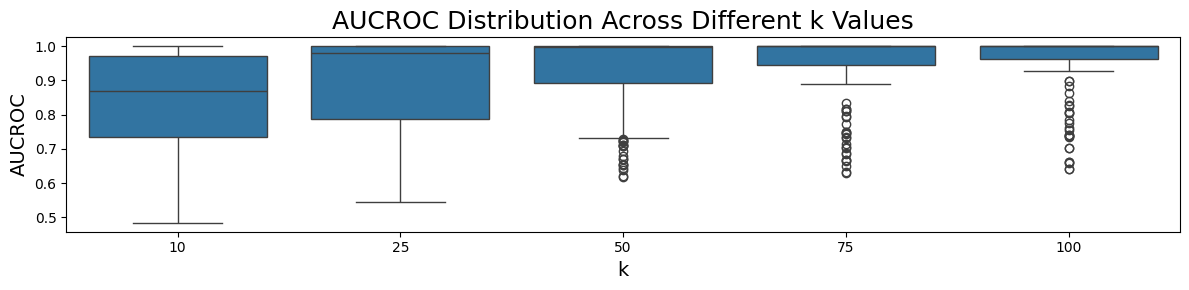}
        \caption{AUC-ROC}
        \label{fig:aucroc_increasing_k}
    \end{subfigure}

    \vspace{0.8em}

    \begin{subfigure}{\textwidth}
        \centering
        \includegraphics[width=1\linewidth, height=4.8cm, keepaspectratio]{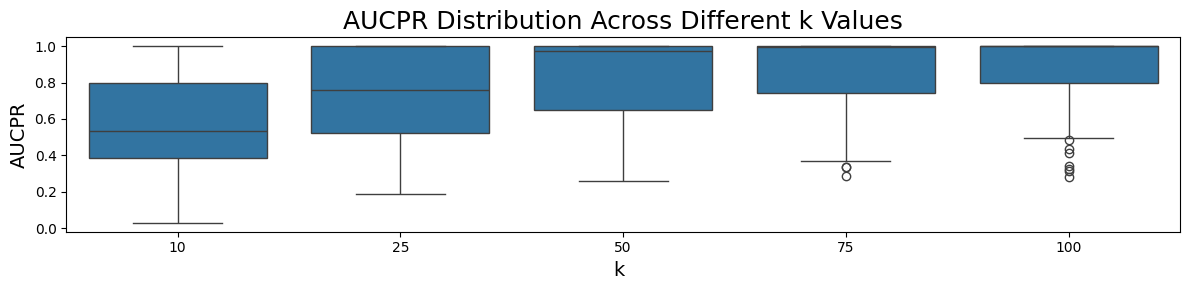}
        \caption{AUC-PR}
        \label{fig:aucpr_increasing_k}
    \end{subfigure}

    \vspace{0.8em}

    \begin{subfigure}{\textwidth}
        \centering
        \includegraphics[width=1\linewidth, height=4.8cm, keepaspectratio]{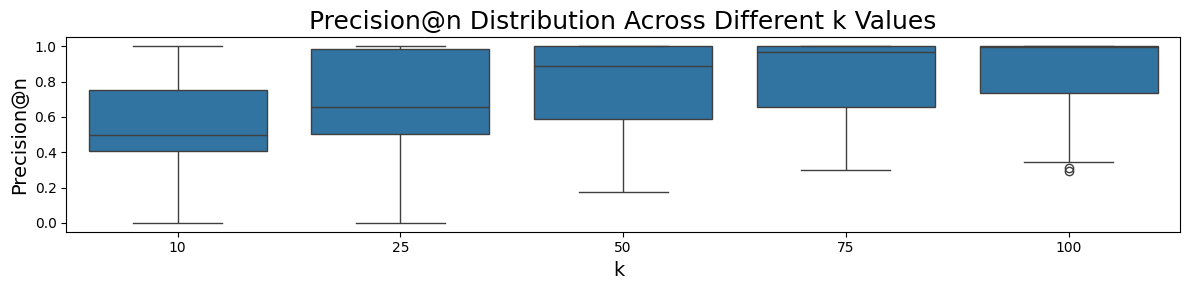}
        \caption{Precision@N}
        \label{fig:precision_increasing_k}
    \end{subfigure}

    \caption{Effect of varying the neighborhood size \(k\) on AUC-ROC, AUC-PR, and Precision@n under the \texttt{realistic\_synthetic\_mode = none} anomaly mode.}
    \label{fig:k_sensitivity}
\end{figure}

This effect is expected for cluster-based anomalies, where anomalous samples occur in spatially coherent groups rather than as isolated points. As illustrated in Figure~\ref{fig:k_sensitivity} and quantified in Table~\ref{tab:k_sensitivity}, increasing the neighborhood size \(k\) systematically enhances the quality of density estimation and mean-shift updates within MSDE. Larger neighborhoods allow each sample to aggregate information from a broader local context, enabling more accurate capture of the collective structure and displacement patterns characteristic of clustered anomalies. This leads to more stable shift directions and improved separation between normal and anomalous regions. For small values of \(k\), the distributions shown in Figure~\ref{fig:k_sensitivity} exhibit higher variability, and the corresponding summary statistics in Table~\ref{tab:k_sensitivity} reflect reduced performance, indicating that overly local neighborhoods are insufficient to reliably represent cluster structure. As \(k\) increases, neighborhood aggregation progressively reveals the underlying anomaly clusters, explaining the consistent performance gains observed across all metrics. The saturation visible in both the figure and the table for \(k \geq 75\) further supports this interpretation, as neighborhoods become large enough to capture the full cluster context, after which additional neighbors contribute limited new information.

\begin{table}[htbp!]
\centering
\caption{Effect of neighborhood size \(k\) on detection performance for cluster anomalies.}
\label{tab:k_sensitivity}
\begin{tabular}{cccc}
\toprule
\textbf{\(k\)} & \textbf{Mean AUC-ROC} & \textbf{Mean AUC-PR} & \textbf{Mean Precision@n} \\
\midrule
10  & 0.842 ± 0.145 & 0.577 ± 0.267 & 0.551 ± 0.257 \\
25  & 0.894 ± 0.136 & 0.734 ± 0.255 & 0.705 ± 0.255 \\
50  & 0.924 ± 0.116 & 0.817 ± 0.227 & 0.790 ± 0.234 \\
75  & 0.942 ± 0.104 & 0.857 ± 0.202 & 0.832 ± 0.214 \\
100 & 0.950 ± 0.096 & 0.883 ± 0.187 & 0.858 ± 0.201 \\
\bottomrule
\end{tabular}
\end{table}

\paragraph{Summary}

Overall, this analysis demonstrates that MSDE performance is predominantly governed by density-related and neighborhood-structure hyperparameters rather than optimization-specific choices.
The hyperparameter importance analysis identifies the neighborhood sample count threshold ($T_{\text{nbd}}$) as the primary driver of performance, while the neighborhood size ($k$ form a secondary but meaningful group of influential parameters.
The dedicated sensitivity analysis of \(k\) under the cluster anomaly mode further corroborates this finding, showing that increasing neighborhood size consistently improves detection performance before saturating at moderate-to-large values.
Taken together, these results indicate that MSDE is robust to the precise tuning of its iterative dynamics, while careful specification of density estimation and neighborhood structure is critical for achieving strong and stable anomaly detection performance.


\section{Additional Experiments under Clean ADBench Setup}
\label{sec:appendix_adbench_clean}

In this section, we report additional experimental results using the ADBench evaluation framework under a clean experimental configuration. Specifically, we consider the setting with \texttt{realistic\_synthetic\_mode = none} and \texttt{noise\_type = none}, which removes both synthetic anomaly injection and artificial noise. This setup allows us to assess the intrinsic anomaly detection capability of each method without external perturbations.

We evaluate the proposed \textbf{MSDE} method alongside standard benchmarking algorithms provided in ADBench, including density-based, proximity-based, linear, and ensemble approaches. All methods are evaluated using identical data splits and preprocessing as defined by the ADBench protocol. All reported scores correspond to the mean performance across datasets.

\subsection{Results and Discussion}
Table~\ref{tab:adbench_clean_results} summarizes the results. MSDE achieves the \textbf{best mean AUC-ROC} among all competing methods, indicating strong global ranking performance in the absence of injected noise. Furthermore, MSDE consistently ranks within the top two methods for both AUC-PR and Precision@N, demonstrating robust detection of true anomalies in highly imbalanced settings.

Interestingly, linear methods such as PCA perform competitively on AUC-PR and P@N in this clean setting, suggesting that low-dimensional structure is informative when noise is absent. However, MSDE maintains strong overall performance across all metrics, highlighting its ability to balance global ranking and early anomaly detection.

\begin{table}[htbp!]
\centering
\caption{Mean anomaly detection performance under clean ADBench setup (\texttt{realistic\_synthetic\_mode = none}, \texttt{noise\_type = none}).}
\label{tab:adbench_clean_results}
\begin{tabular}{lccc}
\toprule
\textbf{Method} & \textbf{AUC-ROC} & \textbf{AUC-PR} & \textbf{P@N} \\
\midrule
MSDE      & \textbf{0.754 ± 0.169} & 0.341 ± 0.277 & 0.319 ± 0.263 \\
IForest   & 0.737 ± 0.173 & 0.338 ± 0.300 & 0.318 ± 0.285 \\
CBLOF     & 0.734 ± 0.184 & 0.339 ± 0.29 & 0.314 ± 0.281 \\
PCA       & 0.728 ± 0.196 & \textbf{0.362 ± 0.297} & \textbf{0.338 ± 0.285} \\
ECOD      & 0.720 ± 0.182 & 0.316 ± 0.258 & 0.293 ± 0.243 \\
COPOD     & 0.719 ± 0.188 & 0.311 ± 0.271 & 0.288 ± 0.262 \\
HBOS      & 0.715 ± 0.184 & 0.322 ± 0.277 & 0.299 ± 0.266 \\
KNN       & 0.707 ± 0.180 & 0.295 ± 0.238 & 0.289 ± 0.227 \\
OCSVM     & 0.694 ± 0.185 & 0.297 ± 0.260 & 0.272 ± 0.242 \\
SOD       & 0.689 ± 0.149 & 0.247 ± 0.202 & 0.263 ± 0.237 \\
LODA      & 0.658 ± 0.210 & 0.276 ± 0.247 & 0.263 ± 0.237 \\
COF       & 0.644 ± 0.160 & 0.221 ± 0.187 & 0.217 ± 0.183 \\
LOF       & 0.638 ± 0.168 & 0.215 ± 0.180 & 0.214 ± 0.178 \\
DAGMM     & 0.629 ± 0.161 & 0.216 ± 0.208 & 0.209 ± 0.203 \\
\bottomrule
\end{tabular}
\end{table}

Overall, these results confirm that MSDE remains highly competitive even in noise-free conditions, achieving strong performance across complementary evaluation metrics. This supports the effectiveness of the proposed density enhancement mechanism beyond settings involving synthetic anomalies or noise injection.

\end{appendices}

\end{document}